\documentclass[times, review, 10pt]{elsarticle}

\usepackage{amssymb}
\usepackage{amsmath}

\usepackage{framed,multirow}
\usepackage{color, colortbl}
\definecolor{Gray}{gray}{0.9}
\definecolor{Gray2}{gray}{0.7}
\definecolor{Gray3}{gray}{0.8}
\definecolor{lightred}{RGB}{230,173,173}
\definecolor{lightgreen}{RGB}{175,230,173}

\usepackage[dvipsnames]{xcolor}
\usepackage{float}
\usepackage[linesnumbered,ruled,vlined]{algorithm2e}% http://ctan.org/pkg/algorithm2e
\DontPrintSemicolon
\usepackage{mdframed}
\usepackage{amssymb}
\usepackage{amsmath}
\usepackage{latexsym}
\usepackage{blindtext}
\usepackage{subcaption}
\usepackage{booktabs}
\usepackage{adjustbox}
\usepackage [autostyle, english = american]{csquotes}
\usepackage{lipsum}
\usepackage{wrapfig}

\usepackage{algorithmic}
\usepackage{xcolor}
\newcolumntype{P}[1]{>{\centering\arraybackslash}p{#1}}

\usepackage{hyperref}

\definecolor{newcolor}{rgb}{.8,.349,.1}

\journal{Pattern Recognition}

\newcommand{\groundtruth}{\mathbf{Y}}
\newcommand{\prediction}{\mathbf{\hat{Y}}}
\newcommand{\inputdata}{\mathbf{X}}

\begin{document}

\begin{frontmatter}

\title{Conformal-in-the-Loop for Learning with Imbalanced Noisy Data}

\author[1]{John Brandon Graham-Knight}
\author[2]{Jamil Fayyad}
\author[3]{Nourhan Bayasi}
\author[1]{Patricia Lasserre}
\author[2]{Homayoun Najjaran\corref{cor1}}
\cortext[cor1]{Corresponding author: \href{homayoun.najjaran@uvic.ca}{homayoun.najjaran@uvic.ca} }

\affiliation[1]{organization={University of British Columbia},
            addressline={}, 
            city={Kelowna},
            postcode={}, 
            state={British Columbia},
            country={Canada}}

\affiliation[2]{organization={University of Victoria},
            addressline={}, 
            city={Victoria},
            postcode={}, 
            state={British Columbia},
            country={Canada}}

\affiliation[3]{organization={University of British Columbia},
            addressline={}, 
            city={Vancouver},
            postcode={}, 
            state={British Columbia},
            country={Canada}}

\begin{abstract}
Class imbalance and label noise are pervasive in large-scale datasets, yet much of machine learning research assumes well-labeled, balanced data, which rarely reflects real-world conditions. Existing approaches typically address either label noise or class imbalance in isolation, leading to suboptimal results when both issues coexist. In this work, we propose Conformal-in-the-Loop (CitL), a novel training framework that addresses both challenges with a conformal prediction-based approach. CitL evaluates sample uncertainty to adjust weights and prune unreliable examples, enhancing model resilience and accuracy with minimal computational cost. Our extensive experiments include a detailed analysis showing how CitL effectively emphasizes impactful data in noisy, imbalanced datasets. Our results show that CitL consistently boosts model performance, achieving up to a 6.1\% increase in classification accuracy and a 5.0 mIoU improvement in segmentation. Our code is publicly available: \href{https://github.com/brandongk-ubco/conformal-in-the-loop}{CitL}.

\end{abstract}

% \begin{graphicalabstract}
% \includegraphics[width=\textwidth]{Picture2.png}
% \end{graphicalabstract}

% \begin{highlights}
% \item We demonstrate that conformal prediction is an effective tool for balancing easy and challenging examples during training.
% \item We design a weighting and pruning mechanism that performs well under label noise, showing particular effectiveness at lower noise levels commonly encountered in practice.
% \item We prune highly uncertain examples at training time by leveraging the Least-Ambiguous set-valued Classifier’s ability to produce empty prediction sets.
% \item We apply dynamic weighting to examples at different stages of training to address class imbalance, with higher weights early on to condition the network on difficult features.
% \item We achieve minimal computational overhead, increasing training time per step by only 11\% for classification and 4\% for segmentation tasks.
% \end{highlights}


\end{frontmatter}

\section{Introduction} \label{sec:intro}
The success of Deep Neural Networks (DNNs) in areas such as image classification~\cite{xie2024described}, object detection~\cite{pu2024rank}, and medical image analysis~\cite{bayasi2024biaspruner,bayasi2024continual} has largely hinged on access to large-scale, accurately labeled datasets. However, producing these high-quality annotations is costly and labor-intensive, especially in fields like medical imaging, where expert annotation is essential. As a result, more cost-effective methods, such as web scraping or crowd-sourcing, have become prevalent~\cite{prabhu2023categories}. While these approaches facilitate data collection, they also introduce noisy labels due to misannotations and ambiguities. When coupled with the strong fitting capacity of DNNs, such label noise can significantly degrade model performance~\cite{song2022learning}. 

Recent approaches to address label noise in datasets generally fall into two categories: Loss/label correction methods~\cite{huang2022contrastive,ghiassi2023trusted}, which aim to adjust noisy labels by estimating a noise transition matrix or using model predictions, or sample selection methods~\cite{xia2023combating,wei2024debiased,zhang2024label}, which identify and select possibly clean data out of noisy ones, and then use them to update the deep networks. Intuitively, if the training data can become less noisy, better generalization can be achieved.

Beyond label noise, another critical challenge in real-world datasets is class imbalance~\cite{dablain2024understanding,aguiar2024survey}. This imbalance arises when certain classes, often referred to as ``head" classes, are represented by a large number of samples, while others, or ``tail" classes, are underrepresented. Datasets with such imbalance tend to bias models toward the majority classes, leading to underrepresentation and under-learning of minority classes. Imbalance-related biases are well-documented: models trained on such data generally optimize for majority-class accuracy, resulting in reduced performance on minority categories. Common approaches to address class imbalance include resampling and reweighting, which adjust either the sample distributions or the class-specific weights to encourage learning from minority classes. 

In many real-world scenarios, noisy labels and class imbalance coexist, presenting a multifaceted problem that single-focus methods struggle to solve. However, techniques aimed at mitigating label noise typically assume a balanced dataset, ignoring the complexities that arise in imbalanced data. Similarly, methods addressing class imbalance generally presume that labels are accurate, failing to account for noise in the underrepresented classes. In such cases, label noise can exacerbate learning bias: when minority classes contain mislabeled samples, models may treat hard-to-learn but clean samples as noise, neglecting genuine examples from the tail classes. This overlap complicates the model’s ability to distinguish between true minority samples and noisy samples from majority classes.

These interconnected challenges highlight the need for a more robust approach to training DNNs on imbalanced and noisy data. To address this, we introduce Conformal-in-the-Loop (CitL), a novel uncertainty-aware training method that dynamically adjusts sample weights and prunes data in response to uncertainty levels calculated through conformal prediction \cite{4410411}. Specifically, CitL applies conformal prediction to assess sample uncertainty, which is then used to strategically modify training in two primary ways: (1) Sample Weighting – assigning higher weights to samples with moderate uncertainty that are likely to be informative, thereby enhancing learning in minority classes, and (2) Sample Pruning – identifying and removing samples with high uncertainty that are likely mislabeled, reducing the noise that would otherwise degrade model performance. By leveraging these two mechanisms, CitL selectively refines the training data, ensuring the model focuses on reliable, informative samples. Through this targeted curation of data, CitL addresses the dual challenges of imbalance and noise while maintaining computational efficiency.  Extensive experimental evaluations demonstrate that CitL significantly improves model performance across common data challenges of imabalanced, noisy labels. To summarize, our contributions in this paper are:

\begin{itemize} 
\item We demonstrate that conformal prediction is an effective tool for balancing easy and challenging examples during training. 
\item We design a weighting and pruning mechanism that performs well under label noise, showing particular effectiveness at lower noise levels commonly encountered in practice. 
\item We prune highly uncertain examples at training time by leveraging the Least-Ambiguous set-valued Classifier (LAC)’s ability to produce empty prediction sets.
\item We apply dynamic weighting to examples at different stages of training to address class imbalance, with higher weights early on to condition the network on difficult features. 
\item We evaluate CitL across multiclass classification and semantic segmentation tasks, demonstrating consistent performance improvements in both settings.
\item We achieve minimal computational overhead, increasing training time per step by only 11\% for classification and 4\% for segmentation tasks.
\end{itemize}

\section{Related Works}
\subsection{Learning with Class Imbalance}
Class imbalance is a pervasive issue in many machine learning tasks, where the majority class is overrepresented in training data, often leading models to underperform on minority classes. This imbalance causes models to overpredict the majority class, resulting in frequent misclassification of minority instances, as noted in several studies~\cite{bauder2018empirical, bauder2018effects}. Solutions for addressing class imbalance generally fall into two main categories: data-centric and algorithmic approaches. 

Data-centric methods focus on adjusting the training data distribution. Undersampling the majority class, though effective, risks discarding critical features, which can lead to a loss of valuable information. Clustering-based undersampling has been proposed to mitigate this by removing redundant or noisy samples while retaining informative examples~\cite{yan2022spatial}. Oversampling the minority class can also help~\cite{sun2022robust,han2023global}, but it can increase the risk of overfitting and prolong training. Synthetic data generation techniques, such as SMOTE~\cite{douzas2018improving}, and data augmentation~\cite{zang2021fasa,chen2022imagine} are commonly used to introduce new information, effectively balancing the data without merely duplicating samples. 

On the other hand, algorithmic approaches often modify the loss function to place greater emphasis on learning from minority samples. One prominent method is Focal Loss, which incorporates a focusing parameter, $\gamma$, to prioritize difficult examples, and a per-class weighting factor, $\alpha$, to counteract imbalance~\cite{Lin2017FocalLossDense}. While $\alpha$ is usually determined by the class distribution, it can overlook the variability in sample quality. The focusing parameter is frequently set to 2 but can be fine-tuned to optimize model performance. Recent advances have introduced additional loss functions tailored for imbalanced datasets. For instance, Label-Distribution-Aware Margin (LDAM) loss encourages larger margins for minority classes, aiming for better generalization in imbalanced settings~\cite{cao2019learning}. Another effective approach is the Distribution-Balanced Loss, which has shown significant improvements in multi-label text classification by re-weighting the contributions of positive instances during training~\cite{wu2020distribution}. 

\subsection{Learning with Noisy Labels}
Learning with noisy labels presents a significant challenge in real-world applications, primarily due to the complexities and costs associated with gathering and accurately labeling supervised training data.  Loss re-weighting, which has been explored extensively in deep learning contexts, is a common approach towards learning with noisy labels. Ren et al.~\cite{ren2019learning} identified the difficulty of distinguishing between mislabelled examples and model uncertainties, complicating the identification of errors. To address this, they propose an additional optimization stage that calculates gradients based on a known clean dataset. This method combines gradients from the training minibatch with those from the clean dataset to optimize the learning process. Furthermore, post-hoc loss re-weighting can be performed based on false positive and false negative error rates, typically estimated through cross-validation. Natarajan et al.~\cite{Natarajan2018CostSensitiveLearning} demonstrated the effectiveness of this re-weighting function in binary classification tasks with synthetic label corruption, recovering the original function with commendable accuracy.

Another approach to addressing noisy labels involves curriculum learning, a method to incrementally increase the complexity of training samples, mirroring human learning processes~\cite{Bengio2009CurriculumLearning}. While curriculum learning strategies have been effectively applied across various machine learning tasks, challenges remain regarding the classification of samples from easy to hard and the selection of appropriate pacing functions for introducing more complex data~\cite{Soviany2022CurriculumLearningSurvey}.  A notable advancement in this area is MentorNet~\cite{Jiang2018MentorNetLearningData}, which can be trained to either follow a predefined curriculum or autonomously discover new ones from the data, adjusting the curriculum based on feedback from another network called StudentNet. This collaborative learning approach allows for dynamic control over the timing and focus of training on individual samples, while StudentNet operates independently during testing. MentorNet's ability to parallelize training makes it scalable for large datasets, with its effectiveness verified on substantial benchmarks.

A foundational work in the field of noisy label learning is Confident Learning (CL), a model-agnostic meta-learning algorithm designed to identify and remove noisy data before training. CL operates in two main phases: first, it estimates class-conditional label noise error rates using cross-validation; second, it employs a rank-and-prune strategy to eliminate the most problematic examples based on loss criteria from the cross-validated model~\cite{Chen2019UnderstandingUtilizingDeep}. CL’s approach includes data pruning through co-teaching, which assesses the generalizability of the dataset—if a dataset lacks generalizability, it indicates the presence of incorrect labels. The algorithm has also been applied iteratively iteratively~\cite{northcutt2021confident}. After identifying and removing noisy samples, CL re-weights the loss for each class to account for the eliminated examples and re-trains on the complete dataset~\cite{northcutt2021confident}. CL is predicated on the assumption that while the labels may be incorrect, the input data is valid, allowing for some useful information to remain as long as the class-conditional probability of noise, $\eta$, is less than or equal to 0.5. As $\eta$ approaches 0.5, training demands more data and time. In essence, CL reverses the concept of boosting by downweighting or excluding high-loss examples, positing that these difficult instances may distort the final model.

\subsection{Uncertainty Quantification}
Uncertainty quantification is important in machine learning for understanding the variability in model predictions~\cite{fayyad2023out}. This variability comes from two main types of uncertainty: epistemic and aleatoric. Epistemic uncertainty arises from a model's limited knowledge and can be reduced by adding more data or improving the model. In contrast, aleatoric uncertainty is the inherent randomness in the data that cannot be reduced. Various methods have been developed to quantify these uncertainties, including Single Deterministic Networks, Bayesian Neural Networks (BNNs), and test-time augmentation techniques. 

Single deterministic networks provide a clear framework for uncertainty estimation using a fixed architecture, allowing predictions through distributions like softmax in classification tasks. However, these networks often struggle with overconfidence in their predictions. Techniques such as Evidential Deep Learning and Prior Networks have been developed to enhance uncertainty quantification~\cite{sensoy2018evidential}. These methods leverage higher-order distributions, with evidential learning predicting parameters of an evidential distribution and prior networks imposing priors on output distributions for better-calibrated uncertainty estimates. In contrast, BNNs address uncertainty by representing model parameters probabilistically, capturing uncertainties through posterior distributions. However, practical implementations often rely on approximations due to the computational intensity of exact Bayesian inference. Monte Carlo Dropout (MCD)~\cite{Gal2016Dropout} is one such approximation, using dropout during inference to sample from the predictive distribution. While averaging multiple predictions provides uncertainty estimates, this requires numerous forward passes, making it time-consuming for large networks. Additionally, selecting appropriate prior distributions for model parameters is critical, as incorrect choices can lead to miscalibrated uncertainty estimates. Similarly, test-time data augmentation generates multiple augmented samples to approximate predictive distributions, also incurring computational overhead. Thus, the effectiveness of these methods often involves a trade-off between uncertainty quantification quality and computational efficiency.

Recently, Conformal Prediction (ConP) has emerged as a robust framework for uncertainty quantification, offering confidence measures around predictions, which are particularly valuable for applications requiring reliable and interpretable outputs. Unlike traditional methods that often rely on assumptions about data distribution or specific model architectures, ConP is model-agnostic and distribution-free. This versatility allows it to be applied to any pre-trained model, whether it be a conventional machine learning algorithm or a complex deep neural network. Consequently, it is suitable for both classification and regression tasks across diverse fields ~\cite{fayyad2024empirical}.

\subsection{Our Contribution}
Our work begins with the observation that the issues of noisy labels and class imbalance often conflict. Class imbalance necessitates that the learning algorithm focuses on hard examples, whereas noisy labels can result in these hard examples being mislabeled. We aim to tackle the challenges of training with noisy, imbalanced data by identifying hard examples for higher weighting while excluding really hard examples that are possibly mislabeled. To achieve this, we apply ConP to enable online data curation in a single training run with minimal overhead. While previous research has leveraged uncertainty quantification to enhance semi-supervised learning—growing training sets through online labeling of unknown examples or pruning out-of-sample instances—these methods often incur high computational costs, typically requiring multiple training runs and cross-validation of models. In contrast, our approach utilizes softmax outputs to generate minimal prediction sets, providing a practical measure of uncertainty. These prediction sets can effectively exclude potentially mislabeled data by yielding empty predictions, streamlining the training process while addressing both class imbalance and noisy labels.

\begin{figure*}
    \centering
    \includegraphics[width=0.8\linewidth]{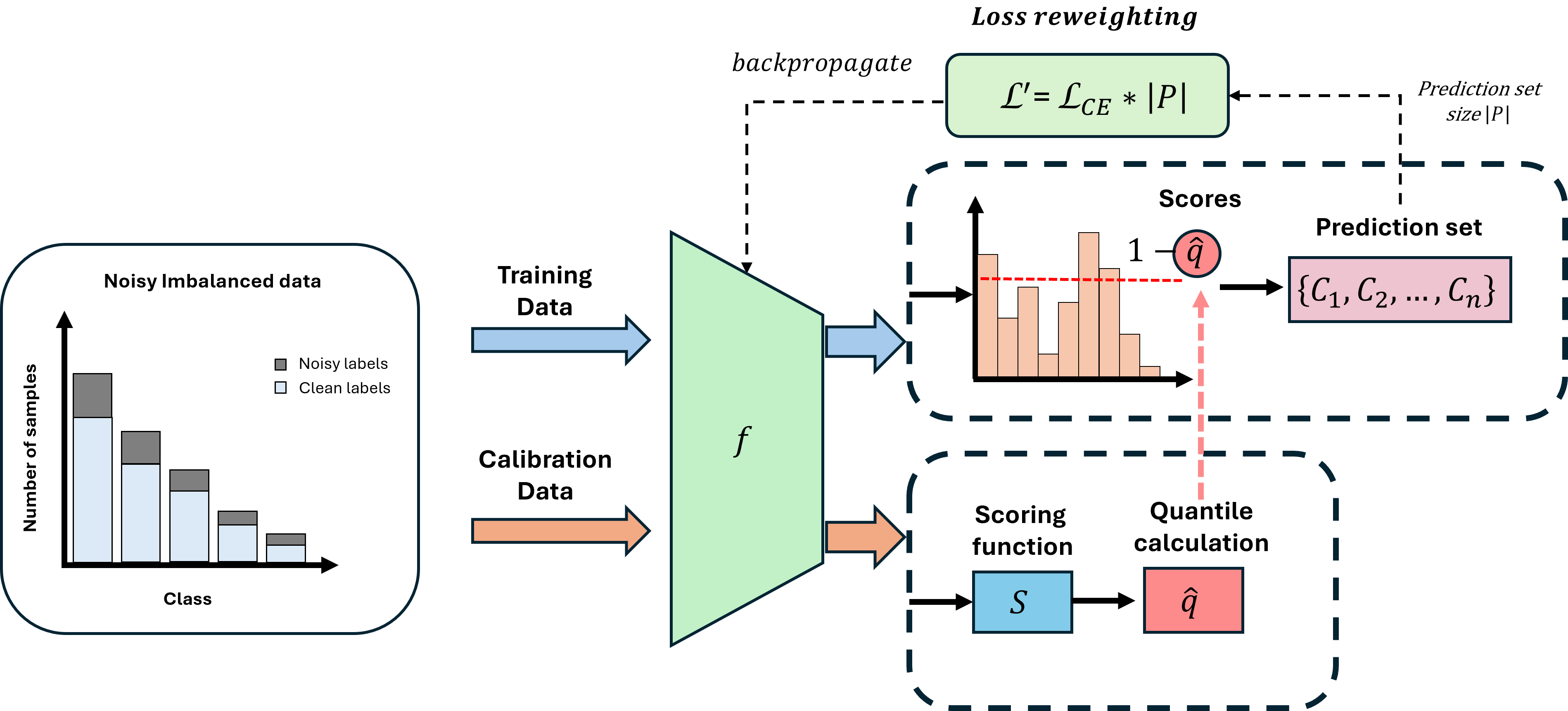}
    \caption{An overview of the proposed Conformal-in-the-Loop (CitL) framework. Conformal Prediction is used to generate prediction sets during self-supervised machine learning. Training examples are weighted by uncertainty; validation examples are used to calibrate the uncertainty model.}
    \label{conformal-in-the-loop}
\end{figure*}

\section{Methodology} \label{sec:Proposed_method}

Conformal Prediction (ConP) is a statistical method that leverages a pre-trained model, such as a neural network classifier, to construct prediction sets with statistically valid uncertainty estimates. The model is first used to generate class probabilities, and a subset of unseen data, designated as calibration data, is used to calibrate the prediction sets. ConP provides a prediction set for each instance such that the probability of containing the correct label is approximately \(1 - \alpha\), where \(\alpha\) is a user-defined error tolerance. This calibration step involves a non-conformity score that quantifies the model’s confidence in its predictions, with higher scores indicating less confidence. This framework ensures statistically valid prediction sets regardless of the underlying model or data distribution.

ConP is widely applicable across fields like computer vision and natural language processing, converting heuristic uncertainties into statistically grounded measures. For a prediction set \(\mathbf{P}\), ConP constructs this set to satisfy:
\begin{equation}
    1 - \alpha \leq P(\hat{Y} \in \mathbf{P}) \leq 1 - \alpha + \frac{1}{n + 1},
\end{equation}
where \( n \) is the number of calibration samples. This bound provides a probabilistic guarantee on the prediction set’s accuracy.

The ConP process involves three primary steps:
\begin{enumerate}
\item \textbf{Non-Conformity Scoring}: Define a heuristic-based non-conformity score function \( s(x, y) \in \mathbb{R} \), where higher scores denote higher uncertainty that sample \( x \) belongs to class \( y \).
\item \textbf{Quantile Calculation}: Calculate the quantile \(\hat{q}\) as the \(\frac{(n+1)(1-\alpha)}{n}\) quantile of the calibration scores \((s(X_1, Y_1), ..., s(X_n, Y_n))\), which adjusts for the desired confidence level.
\item \textbf{Prediction Set Construction}: Formulate the prediction set for new instances as \( C(X_{\text{test}}) = \{y : s(X_{\text{test}}, y) \leq \hat{q}\} \).
\end{enumerate}

Two common choices for non-conformity scoring are the Least Ambiguous Classifier (LAC) and the Adaptive Prediction Set (APS). LAC, represented by \(1 - \prediction_c\), where \(\prediction_c\) is the model’s output probability for the correct label, is computationally efficient and ensures marginal coverage guarantees~\cite{Sadinle_2018}. APS, in contrast, sums the model outputs that are higher than the score of the true class:
\begin{equation}
    s(\inputdata, \groundtruth) = \sum_{j=1}^k f(\inputdata_i) \pi_j,
\end{equation}
\noindent where \(\prediction_y = \pi_k\) for the ground truth class \(k\). APS iteratively includes classes in the prediction set until the cumulative probability meets the target quantile~\cite{romano2020classification}. While LAC may result in empty prediction sets if no scores exceed the target quantile, APS always produces a non-empty set.

Recent research highlights ConP’s computational efficiency and theoretical validity for training-phase applications. For instance,~\cite{einbinder2022training} introduces a differentiable approximation of the cumulative distribution function (CDF) of softmax outputs, which informs a novel loss function that incorporates calibration uncertainty into the training process. This method enables more compact prediction sets and enhances conditional coverage through precise calibration, supported by a hold-out calibration dataset.

\subsection{Proposed CitL Method}
Fig.~\ref{conformal-in-the-loop} provides an overview of the proposed Conformal-in-the-Loop (CitL) framework. At a high level, CitL integrates conformal prediction (ConP) into the training process to adjust the weighting of examples, thereby enhancing overall network performance.  Formally, given a set of examples $\{ \mathbf{X}_1, \dots, \mathbf{X}_n \}$ and ground truth labels $\{ Y_1, \dots, Y_n \}$, we use a neural network to learn an approximate classification function $f : \mathbf{X} \rightarrow \hat{Y}$. The classifier is accurate when the function produces the correct ground truth class, $\hat{Y} = Y$. However, using ConP, we allow $\hat{Y}$ to take on a set of class labels from possible classes $\{ C_1, \dots, C_m \}$, rather than a single label, which we call the prediction set $\mathbf{P}$. We begin with the notion that the size of the prediction set is an estimate of the practical uncertainty of a model.  Given the possibility of empty prediction sets, both pruning and weighting are achieved by multiplying the loss by the prediction set size:
\begin{equation}
    \mathcal{L}'_x = \lvert \mathbf{P}_x \rvert ~ \mathcal{L}_x.
\end{equation}

Notably, the weighting of the loss is not required to be differentiable, removing the greatest limitation to applying additional non-conformity scores, as experienced in~\cite{einbinder2022training}. 

Since ConP can be achieved using only the softmax outputs of the classifier, which are required for training already, we can cache the non-conformity scores from each batch and pass them to the conformal classifier. We summarize our epoch training process in Algorithm~\ref{alg:1}. Note that this is a slightly modified standard iterative mini-batch optimization algorithm, with line 4 added to calculate the conformal prediction sets and line 5 modified to weight the loss by prediction set size. The validation epoch is also modified to fit the conformal classifier on the first portion of the dataset, which then evaluates the uncertainty on the remaining data, as summarized in Algorithm~\ref{alg:2}.

\begin{algorithm}[t]
\caption{CitL Training Epoch}
\label{alg:citl_training}

\begin{algorithmic}[1]

\REQUIRE{A parameterized function $f'(\mathbf{X}, \mathbf{W})$}
\REQUIRE{A fit conformal classifier C}
\REQUIRE{A loss function $\mathcal{L}(\hat{Y}, Y)$}

\WHILE{Training Data is not Exhausted}
    \STATE $\mathbf{\beta} \gets \text{Random Sample from } \{ (\mathbf{X}_0, Y_0), ... (\mathbf{X}_n, Y_n) \}$
    \STATE $\beta_{\hat{Y}} \gets f'(\mathbf{\beta_x}, \mathbf{W})$
    \STATE $\beta_\mathbf{P} \gets C(\beta_{\hat{y}})$
\STATE $\mathbf{W} \gets \mathbf{W} - \eta \nabla (\lvert \beta_\mathbf{P} \rvert \mathcal{L}(\beta_{\hat{Y}}, \beta_Y))$

\ENDWHILE

\end{algorithmic}
\label{alg:1}
\end{algorithm}

\begin{algorithm}[t]
\caption{CitL Validation Epoch}
\label{alg:citl_validation}
\begin{algorithmic}[1]

\REQUIRE{A parameterized function $f'(\mathbf{X}, \mathbf{W})$}
\REQUIRE{A loss function $\mathcal{L}(\hat{Y}, Y)$}
\REQUIRE{Calibration Set Size $\lvert A \rvert$}

\STATE $V_x \gets []$
\STATE $V_{\hat{Y}} \gets []$
\STATE $V_Y \gets []$
\WHILE{Validation Data is not exhausted}
    \STATE $\mathbf{\beta} \gets \text{Random Sample from } \{ (\mathbf{X}_0, Y_0), \dots, (\mathbf{X}_n, Y_n) \}$
    \STATE $\beta_{\hat{Y}} \gets f'(\mathbf{\beta_x}, \mathbf{W})$
    \IF{idx $< \lvert A \rvert$}
        \STATE $V_x$.append$(\mathbf{\beta_x})$
        \STATE $V_{\hat{Y}}$.append$(\beta_{\hat{Y}})$
        \STATE $V_Y$.append$(\beta_Y)$
    \ELSIF{idx $= \lvert A \rvert$}
        \STATE Fit C from $V_x$, $V_{\hat{Y}}$, and $V_Y$.
    \ELSE
        \STATE $\beta_\mathbf{P} \gets C(\beta_{\hat{Y}})$
    \ENDIF
\ENDWHILE

\end{algorithmic}
\label{alg:2}
\end{algorithm}

Fitting the conformal classifier introduces memory and computational complexity proportional to the size of the calibration set, making it advantageous to keep this set as small as possible while still ensuring representative sampling of the data. During training, and once the conformal classifier is calibrated in the validation step, computational dependency shifts to the mini-batch size, which is typically manageable in most setups.

%%%%%%%%%%%%%%%%%%%%%%%%%%%%%%%%%%%%%%%%%%%%%%%%%%%%%%%%%%%%%%%%%%%
\subsection{Evaluation Benchmarks}
For our evaluation benchmarks, we focus on two tasks: multiclass classification and semantic segmentation. For multiclass classification, we select CIFAR-10since it is known to be well-balanced; we then synthetically introduce label noise and class imbalance.  For semantic segmentation, we select the CityScapes dataset, which is known to be imbalanced.

\noindent \textbf{Multiclass Classification.} We utilize the CIFAR-10 dataset, which consists of 32x32 color images. Given the small dimensions of these images, we resize them to 224x224x3 to provide adequate training information for the network. The dataset includes 10 classes, each with 6000 images, totaling 60,000 images. The validation set is created by randomly sampling 10\% of the training examples. 

\noindent \textbf{Semantic Segmentation.} We use the CityScapes dataset, which contains 2048x1024 RGB images of urban street scenes. Training images are patched to 512x1024x3, effectively reducing the total number of pixels by a factor of four to allow a batch size of 6. Validation and testing are conducted at full resolution with a batch size of 1.  Class labels are mapped to the 20-class labelling method recommended in the CityScapes documentation~\cite{Cordts2016CityScapesDatasetSemantic}, and the background class is ignored when calculating loss or evaluation metrics.  Since the test dataset provided by CityScapes has no public annotations, we reserve 20\% of the 3,477 image train set for validation; the 500 image val set is reserved for our test set.

\subsection{Experimental Setup and Implementation Details}
\noindent \textbf{Multiclass Classification with CIFAR-10. } To examine the method's capability to identify and mitigate class imbalance, we randomly downsample the training and validation sets for the Airplane and Automobile classes to 20\% of the total available examples for each class, with this downsampling performed prior to the train-validation split and not applied to the test set. Label noise is synthetically added to the remaining training examples; importantly, no label noise is introduced into the validation or test sets to maintain accurate labels for this small subset. For calibration, we allocate 20\% of the validation images from CIFAR-10 as the calibration set, enabling us to utilize a larger portion of the validation set for calibration.

\noindent \textbf{Semantic Segmentation with CityScapes.} We allocate 10\% of the validation images as the calibration set. Given the large number of pixels involved in semantic segmentation, quantile estimation can be computationally intensive; thus, we uniformly sample 10\% of the pixel examples from the calibration set, resulting in a total of 1\% of the validation pixels being used for fitting the quantile.

\noindent \textbf{Other Details.} We implement our method, ConP, using the Least Ambiguous Set-Valued Classifier (LAC)~\cite{Sadinle_2018} in PyTorch, leveraging GPU acceleration for enhanced computational efficiency. Data augmentation is performed using the Albumentations library, primarily aimed at preventing model overfitting. For CIFAR-10, we apply augmentations such as Horizontal Flip (50\%), Crop and Pad (ranging from 0 to 56 pixels), and Coarse Dropout (covering a 56x56 area). For the CityScapes dataset, augmentation begins with a random scaling of the image size by up to ±10\%, followed by a potential rotation of up to 10 degrees occurring 50\% of the time. The image is then randomly cropped to a fixed size of 512x1024 pixels, with a 50\% probability of horizontal flipping, and there is a 25\% chance of applying one of several transformations, including reducing image quality to 90\%, applying a blur, or adding Gaussian noise, along with brightness and contrast adjustments performed in 25\% of cases.

\subsection{Training Setup}
The training environment includes a devcontainer with Python v3.12, PyTorch Lightning v2.4.0, PyTorch v2.4, and CUDA 12.1.0. Classification tasks are performed using MNasNet-Small from the timm library, while the segmentation task employs EfficientNet-b0 in a U-Net configuration from Segmentation Models PyTorch. Classification models are trained from scratch, whereas segmentation models use transfer learning from pretrained ImageNet weights.  The best model to evaluate against the test set is selecting using minimum validation loss for baseline models.  When using the method, the dynamic loss reweighting means that loss values cannot be effectively compared against epochs, so we select the best model by maximum validation accuracy for classification and mIoU for segmentation.

The learning rate is fixed at $5 \times 10^{-4}$ and is reduced by a factor of 0.2 upon plateau, with a patience of 10 epochs. The minimum learning rate is set to $10^{-6}$, and weight decay is configured at 10\% of the initial learning rate. Early stopping is applied based on validation loss with a patience of 20 epochs. The ADAM optimizer is used with default parameters $\beta = (0.9, 0.999)$. All experiments are conducted on a single A6000 GPU with 256GB of system RAM.

For segmentation, weighting is unnormalized, allowing examples to be weighted anywhere from 0 to the total number of classes, which is 19 in the case of CityScapes.  Although this was found to work well with the smooth gradient provided by the large number of examples in segmentation, classification has a much smaller number of examples per batch, leading to large variations in gradient magnitude.  To mitigate the effect of this variation on learning rate, the weighting of each batch is normalized by the mean weight value in the batch.

We conduct a grid search to identify the optimal \(\alpha\) values, exploring a range of 0.10 to 0.19 for classification and 0.01 to 0.10 for segmentation tasks.

\subsection{Evaluation Metrics}
Given \( N \) as the number of classes, \( TP_i \) as the number of true positives for class \( i \), \( TN_i \) as the number of true negatives for class \( i \), \( FP_i \) as the number of false positives for class \( i \), and \( FN_i \) as the number of false negatives for class \( i \),  we compute the macro accuracy for classification tasks as follows:

\begin{equation}
    \text{Accuracy} = \frac{1}{N} \sum_{i=1}^{N} \frac{TP_i + TN_i}{TP_i + TN_i + FP_i + FN_i}.
\end{equation}

This metric provides an average accuracy across all classes, allowing for a balanced assessment of model performance, particularly in scenarios involving class imbalance. For segmentation tasks, we evaluate performance using the mean Intersection over Union (mIoU), defined as:

\begin{equation}
\text{mIoU} = \frac{1}{N} \sum_{i=1}^{N} \frac{TP_i}{TP_i + FP_i + FN_i}.
\end{equation}

To address potential issues with undefined mIoU values—especially when a class is not represented and the union is zero—we concatenate all examples for a given class in the test set before performing the calculation.

\subsection{Reproducibility}
The complete implementation is available on \href{https://github.com/brandongk-ubco/conformal-in-the-loop}{Github}. Training runs and ONNX models are recorded and available in \href{https://app.neptune.ai/o/conformal-in-the-loop/org/citl/runs}{Neptune.ai}.

% \newpage

\section{Results and Discussion}

\begin{figure}[ht]
    \centering
    \begin{minipage}{0.45\textwidth}
    \centering
    \includegraphics[width=\textwidth]{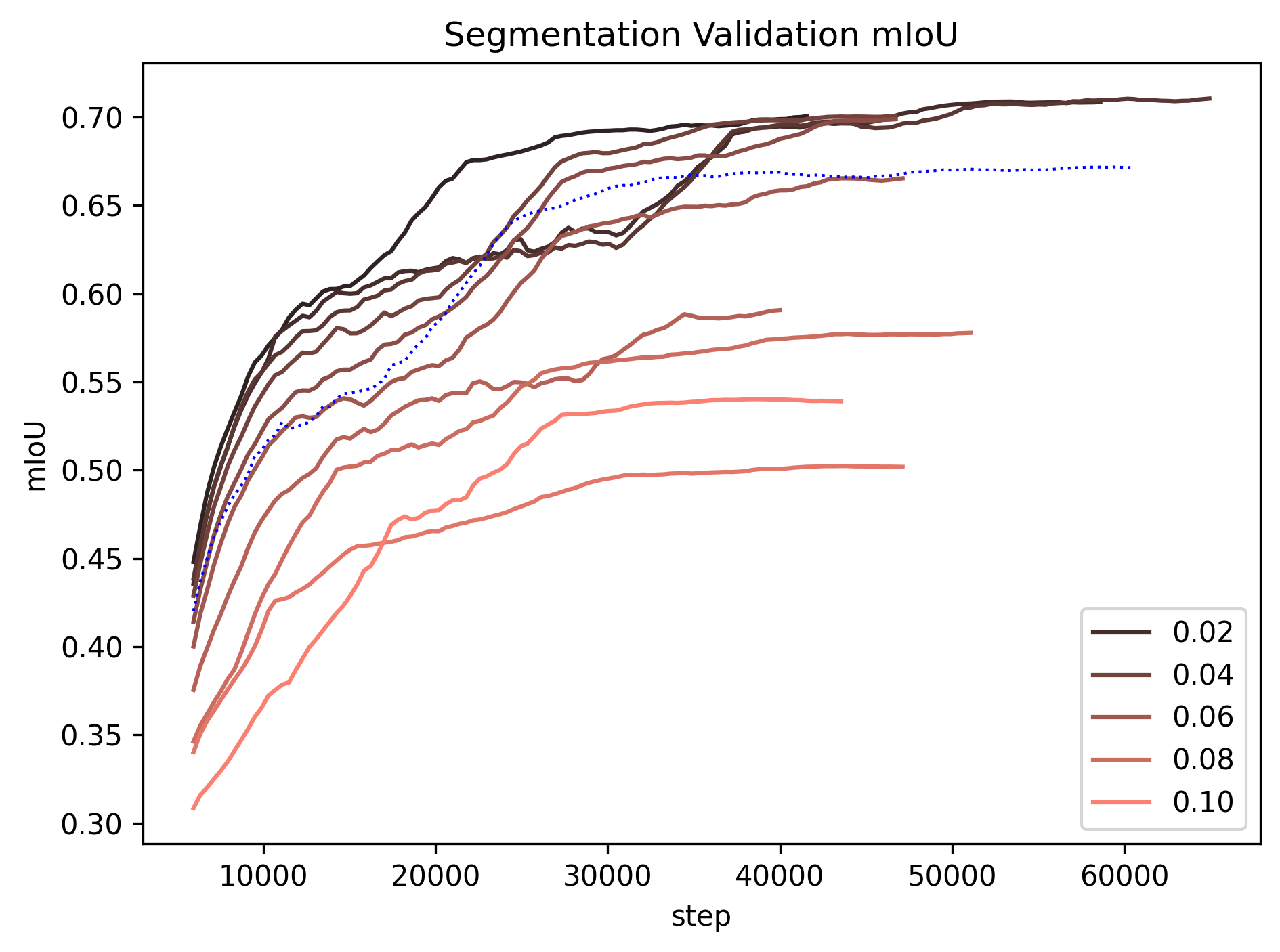}
    \caption{Segmentation mIoU on the validation set for various values of $\alpha$. The baseline performance is indicated by the dotted blue line.}
    \label{fig:segmentation_val_miou}
    \end{minipage}\hfill
    \begin{minipage}{0.45\textwidth}
    \centering
    \includegraphics[width=\columnwidth]{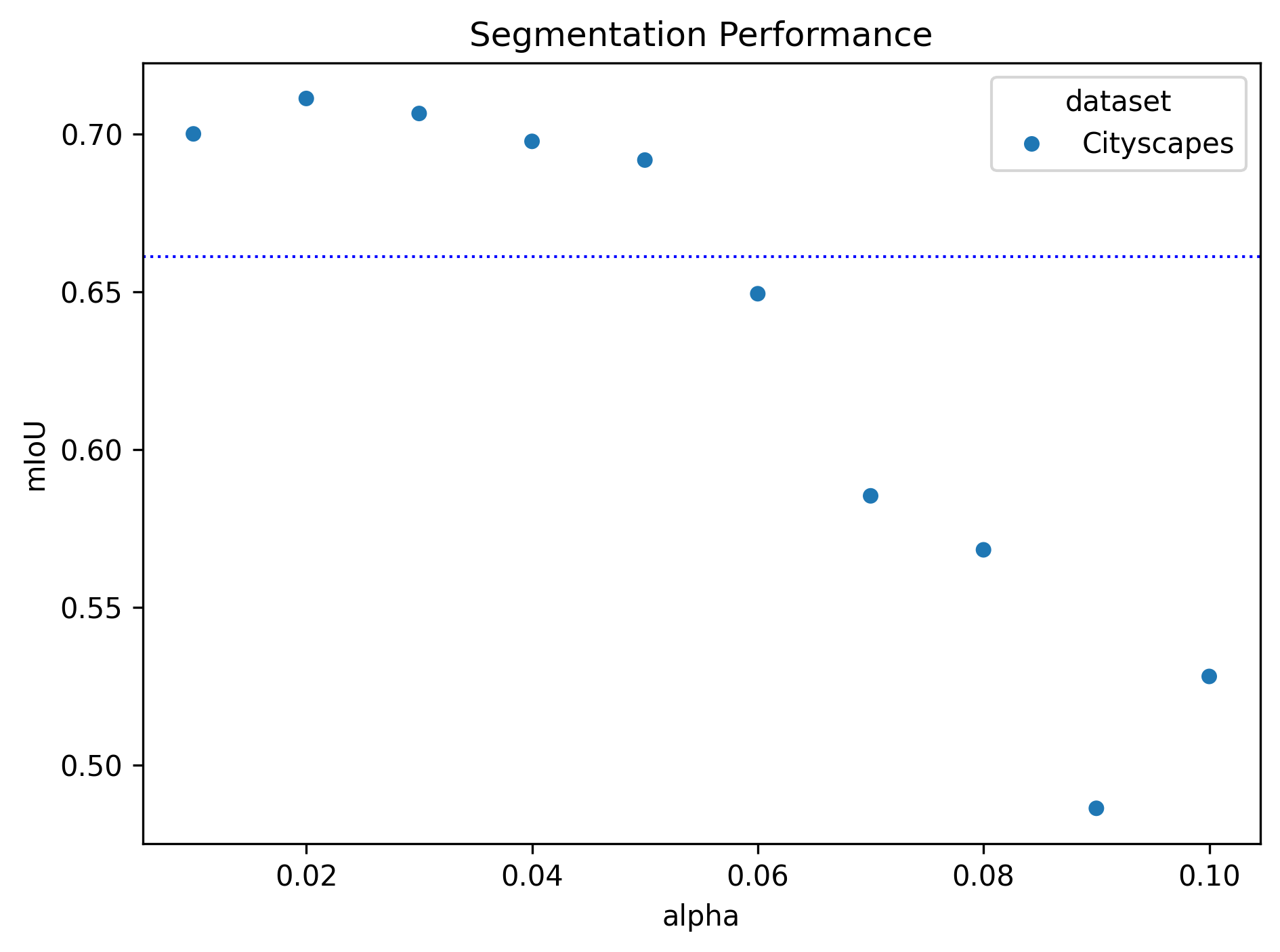}
    \caption{Segmentation mIoU for values of $\alpha$ [0.01, 0.10]. The best mIoU of 71.1 is achieved at $\alpha = 0.02$. Baseline mIoU (66.1) is represented by the dotted blue line.}
    \label{fig:segmentation_results}
    \end{minipage}
\end{figure}

We compare the proposed CitL method against a baseline model trained with either conventional cross-entropy loss or focal loss. Unless specified otherwise, the term `baseline' refers to the model trained using cross-entropy loss. Furthurmore, we conduct six distinct experiments for classification, each varying the noise level from 0\% to 50\%. For segmentation, we perform a single experiment at 0\% noise.

\subsection{Preliminary Experiment: Selecting the \(\alpha\) Value}
From the grid search, we identify the optimal \(\alpha\) values for each experiment. Table~\ref{tab:best_alpha_metrics} summarizes the best performance metrics—accuracy for CIFAR-10 and mIoU for CityScapes—achieved by our CitL method with these optimal \(\alpha\) values, as well as the performance of the baseline model. These identified \(\alpha\) values are consistently utilized in subsequent analyses across all corresponding experiments.

\begin{table}[t]
\centering
\small
\caption{Performance metrics (accuracy for CIFAR-10 and mIoU for CityScapes) for the CitL method with optimal \(\alpha\) values compared to the baseline model across different datasets. \(\Delta\) represents the difference in performance between CitL and the baseline. }
\label{tab:best_alpha_metrics}
\begin{tabular}{rc|c|rr|r}
\toprule
 & Noise \% & $\alpha$ & Baseline & Method & $\Delta$ \\
\midrule
\multirow{6}{*}{\shortstack{Imbalanced \\ CIFAR-10}} & 0 & 0.18 & 83.0 & 85.5 & 2.5 \\
 & 10 & 0.11 & 75.5 & 79.5 & 4.0 \\
 & 20 & 0.14 & 74.0 & 77.1 & 3.1 \\
 & 30 & 0.13 & 69.4 & 72.5 & 3.1 \\
 & 40 & 0.11 & 67.4 & 70.8 & 3.4 \\
 & 50 & 0.17 & 63.3 & 69.4 & 6.1 \\
\cline{1-6} \cline{2-6}
CityScapes & - & 0.02 & 66.1 & 71.1 & 5.0 \\
\cline{1-6} \cline{2-6}
\bottomrule
\end{tabular}
\end{table}

\begin{table*}[t]
\centering
\caption{Classwise Accuracy for the imbalanced CIFAR-10 dataset across various levels of synthetic noise. The table shows the baseline performance, performance with the method applied, and the resulting improvement for each class.  Light red represents the worst-performing class (airplane in all cases) whose accuracy is improved by the method at all noise levels.  Light green represents the best-performing class (ship or truck).}
\label{tab:classwise_cifar10}
\adjustbox{max width=\textwidth}{%
\begin{tabular}{r||rrr|rrr|rrr|rrr|rrr|rrr}
\toprule
\multicolumn{1}{r}{Noise Level} & \multicolumn{3}{c|}{0\%} & \multicolumn{3}{c|}{10\%} & \multicolumn{3}{c|}{20\%} & \multicolumn{3}{c|}{30\%} & \multicolumn{3}{c|}{40\%} & \multicolumn{3}{c}{50\%} \\
\multicolumn{1}{r}{} & baseline & method & $\Delta$ & baseline & method & $\Delta$ & baseline & method & $\Delta$ & baseline & method & $\Delta$ & baseline & method & $\Delta$ & baseline & method & $\Delta$ \\
\midrule
Airplane & \cellcolor{lightred} 64.4 & \cellcolor{lightred} 73.6 & 9.2 & \cellcolor{lightred} 57.4 & \cellcolor{lightred} 62.0 & 4.6 & \cellcolor{lightred} 45.3 & \cellcolor{lightred} 62.4 & 17.1 & \cellcolor{lightred} 42.1 & \cellcolor{lightred} 48.2 & 6.1 & \cellcolor{lightred} 45.0 & \cellcolor{lightred} 49.1 & 4.1 & \cellcolor{lightred} 34.4 & \cellcolor{lightred} 52.7 & 18.3 \\
Automobile & 83.8 & 82.0 & -1.8 & 74.4 & 80.4 & 6.0 & 72.4 & 83.3 & 10.9 & 59.2 & 74.3 & 15.1 & 51.4 & 66.8 & 15.4 & 57.4 & 72.7 & 15.3 \\
Bird & 81.9 & 83.5 & 1.6 & 70.3 & 77.4 & 7.1 & 74.6 & 72.8 & -1.8 & 74.9 & 68.2 & -6.7 & 61.7 & 66.8 & 5.1 & 62.0 & 63.1 & 1.1 \\
Cat & 73.9 & 76.3 & 2.4 & 64.2 & 66.7 & 2.5 & 60.5 & 66.1 & 5.6 & 39.6 & 51.6 & 12.0 & 53.3 & 61.5 & 8.2 & 47.4 & 48.5 & 1.1 \\
Deer & 83.8 & 88.9 & 5.1 & 78.0 & 82.5 & 4.5 & 70.4 & 77.2 & 6.8 & 63.8 & 69.6 & 5.8 & 59.2 & 69.9 & 10.7 & 67.5 & 73.0 & 5.5 \\
Dog & 76.2 & 81.3 & 5.1 & 68.2 & 73.1 & 4.9 & 69.9 & 72.7 & 2.8 & 74.6 & 75.0 & 0.4 & 72.1 & 65.9 & -6.2 & 62.5 & 65.1 & 2.6 \\
Frog & 89.2 & 89.0 & -0.2 & 83.2 & 84.4 & 1.2 & 86.0 & 80.7 & -5.3 & 83.0 & 81.4 & -1.6 & 82.2 & 79.0 & -3.2 & 71.9 & 73.2 & 1.3 \\
Horse & 90.0 & 92.5 & 2.5 & 79.7 & 88.7 & 9.0 & 81.9 & 82.0 & 0.1 & 81.3 & 83.3 & 2.0 & 78.2 & 76.3 & -1.9 & 70.1 & 78.2 & 8.1 \\
Ship & 93.5 & \cellcolor{lightgreen} 94.0 & 0.5 & \cellcolor{lightgreen} 90.2 & \cellcolor{lightgreen} 90.6 & 0.4 & 88.2 & \cellcolor{lightgreen} 88.2 & 0.0 & \cellcolor{lightgreen} 91.6 & \cellcolor{lightgreen} 87.3 & -4.3 & \cellcolor{lightgreen} 85.8 & \cellcolor{lightgreen} 87.7 & 1.9 & 79.0 & \cellcolor{lightgreen} 86.1 & 7.1 \\
Truck & \cellcolor{lightgreen} 93.7 & 93.8 & 0.1 & 89.6 & 89.6 & 0.0 & \cellcolor{lightgreen} 90.5 & 85.5 & -5.0 & 84.1 & 85.6 & 1.5 & 85.1 & 84.9 & -0.2 & \cellcolor{lightgreen} 81.2 & 81.7 & 0.5 \\
\midrule
Best - Worst & 29.3 & 20.4 & -8.9 & 32.8 & 28.6 & -4.2 & 45.2 & 25.8 & -19.4 & 49.5 & 39.1 & -10.4 & 40.8 & 38.6 & -2.2 & 46.8 & 33.4 & -13.4  \\
Overall & 83.0 & 85.5 & 2.4 & 75.5 & 79.5 & 4.0 & 74.0 & 77.1 & 3.1 & 69.4 & 72.5 & 3.0 & 67.4 & 70.8 & 3.4 & 63.3 & 69.4 & 6.1 \\
\bottomrule
\end{tabular}
}
\end{table*}

\subsection{Classification Results}

\begin{figure}[t]
    \centering
    \includegraphics[width=0.48\textwidth]{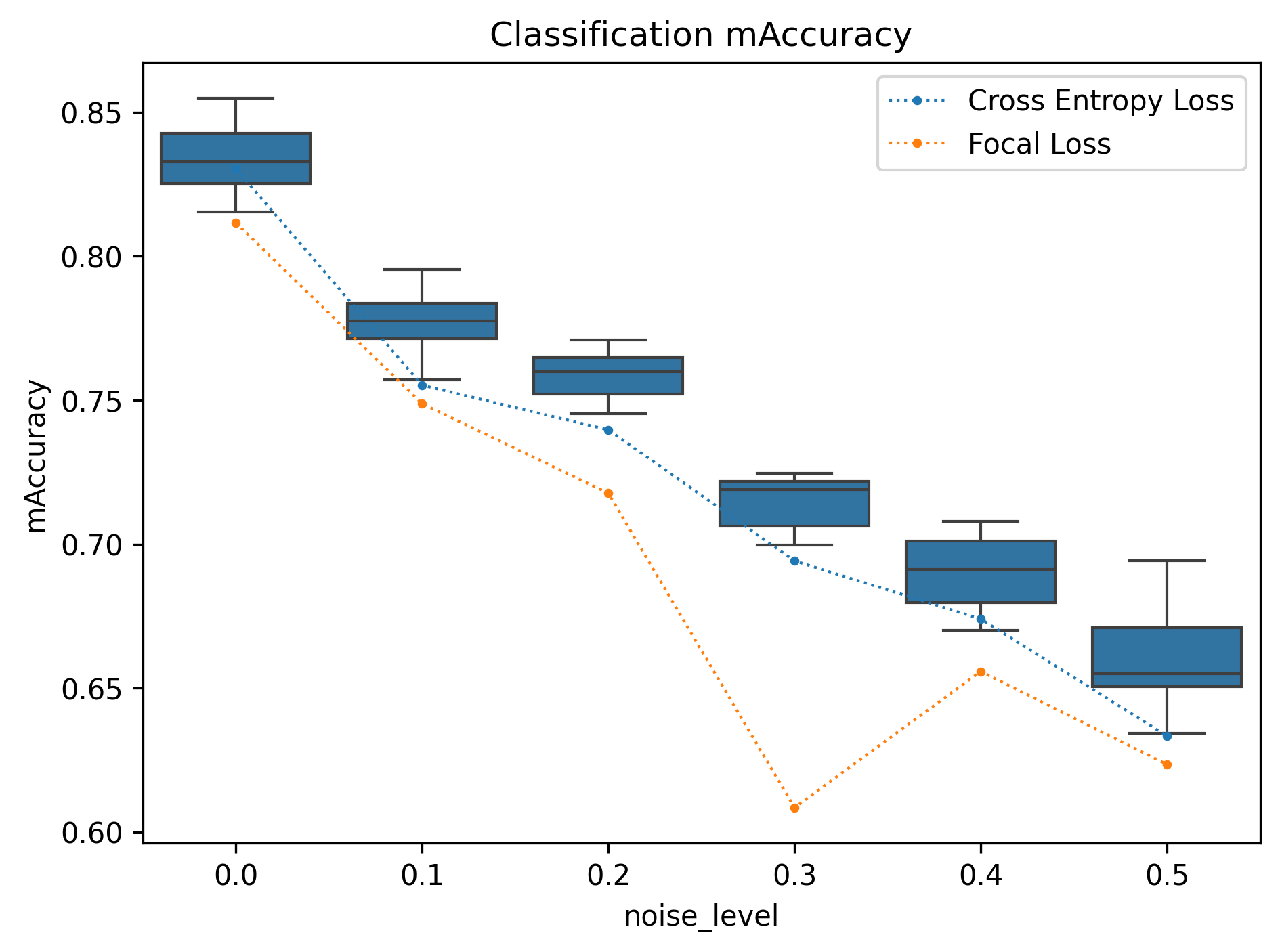}
    \caption{Classification accuracy on imbalanced CIFAR-10 using our CitL method and the baseline with cross-entropy loss (blue dotted line) and focal loss (orange dotted line). The box plots show 10 $\alpha$ values ranging from 0.10 to 0.19.}
    \label{fig:classification_accuracy}
\end{figure}

Fig.~\ref{fig:classification_accuracy} illustrates the accuracy of our method on the CIFAR-10 dataset across varying levels of synthetic noise added to the training set. Notably, our method performs comparably to the cross-entropy baseline when no noise is introduced and surpasses baseline performance at all noise levels above 0\%.

Table~\ref{tab:classwise_cifar10} provides a detailed breakdown of the method's performance using the optimal $\alpha$ value (given in Table~\ref{tab:best_alpha_metrics}) across different noise levels, categorized by class. As expected, the method demonstrates significant improvements in the downsampled classes of airplane and automobile. At every noise level, the airplane class exhibits the lowest performance; therefore, enhancing performance in this class contributes to achieving a more balanced model overall.

\subsection{Segmentation Results}

\begin{wraptable}{r}{0.33\textwidth}
\vspace*{-1cm}
\centering
\caption{Classwise IoU for the CityScapes dataset with the best training run using $\alpha = 0.02$.}
\label{tab:classwise_CityScapes}
\begin{adjustbox}{width=0.30\textwidth}
\begin{tabular}{r|rrr}
\toprule
 & Baseline & Method & $\Delta$ \\
\midrule
Road & 97.3 & 97.5 & 0.2 \\
Sidewalk & 80.3 & 81.3 & 1.0 \\
Building & 90.1 & 90.8 & 0.7 \\
Wall & 45.6 & 45.1 & -0.5 \\
Fence & 47.5 & 51.3 & 3.8 \\
Pole & 58.1 & 60.9 & 2.8 \\
Traffic light & 57.9 & 64.3 & 6.3 \\
Traffic sign & 72.0 & 74.3 & 2.3 \\
Vegetation & 91.3 & 91.8 & 0.5 \\
Terrain & 56.9 & 58.5 & 1.6 \\
Sky & 93.2 & 94.1 & 0.9 \\
Person & 76.7 & 78.8 & 2.2 \\
Rider & 50.9 & 55.0 & 4.1 \\
Car & 92.8 & 93.5 & 0.7 \\
Truck & 49.0 & 66.7 & 17.7 \\
Bus & 48.0 & 66.0 & 18.1 \\
Train & 39.6 & 59.6 & 20.1 \\
Motorcycle & 36.7 & 47.8 & 11.1 \\
Bicycle & 71.9 & 73.8 & 1.9 \\
\midrule
Overall & 66.1 & 71.1 & 5.0 \\
\bottomrule
\end{tabular}
\end{adjustbox}
\end{wraptable}

Fig.~\ref{fig:segmentation_results} illustrates the segmentation results of our method on the CityScapes dataset using various values of \(\alpha\), comparing them to the baseline model. At lower values of \(\alpha\), our method significantly outperforms the baseline, demonstrating a clear advantage in segmentation accuracy. However, as \(\alpha\) increases, we observe a noticeable decline in performance.

Fig.~\ref{fig:segmentation_val_miou} presents the validation mean Intersection over Union (mIoU) across training epochs. The results indicate that improvements in mIoU due to weighting become evident early in the training process, with significant gains observed after just a few epochs. The decline in performance with increasing \(\alpha\) is discussed in more detail in Section~\ref{sec:pruning}.

Table~\ref{tab:classwise_CityScapes} details the performance of our method with \(\alpha = 0.02\) by class. As expected, the method shows notable improvements in challenging classes such as motorcycle, truck, bus, rider, train, fence, traffic light, traffic sign, bicycle, person, terrain, pole, and sidewalk, with four of these classes achieving mIoU improvements exceeding 10. Additionally, even some of the relatively easier classes, including building, vegetation, road, and sky, exhibit minor performance gains when our method is applied. Only the wall class showed a modest decrease in mIoU of 0.5.

\subsection{Computational Overhead}

\begin{figure}[t]
    \begin{minipage}{0.48\textwidth}
    \centering
    \includegraphics[width=1\linewidth]{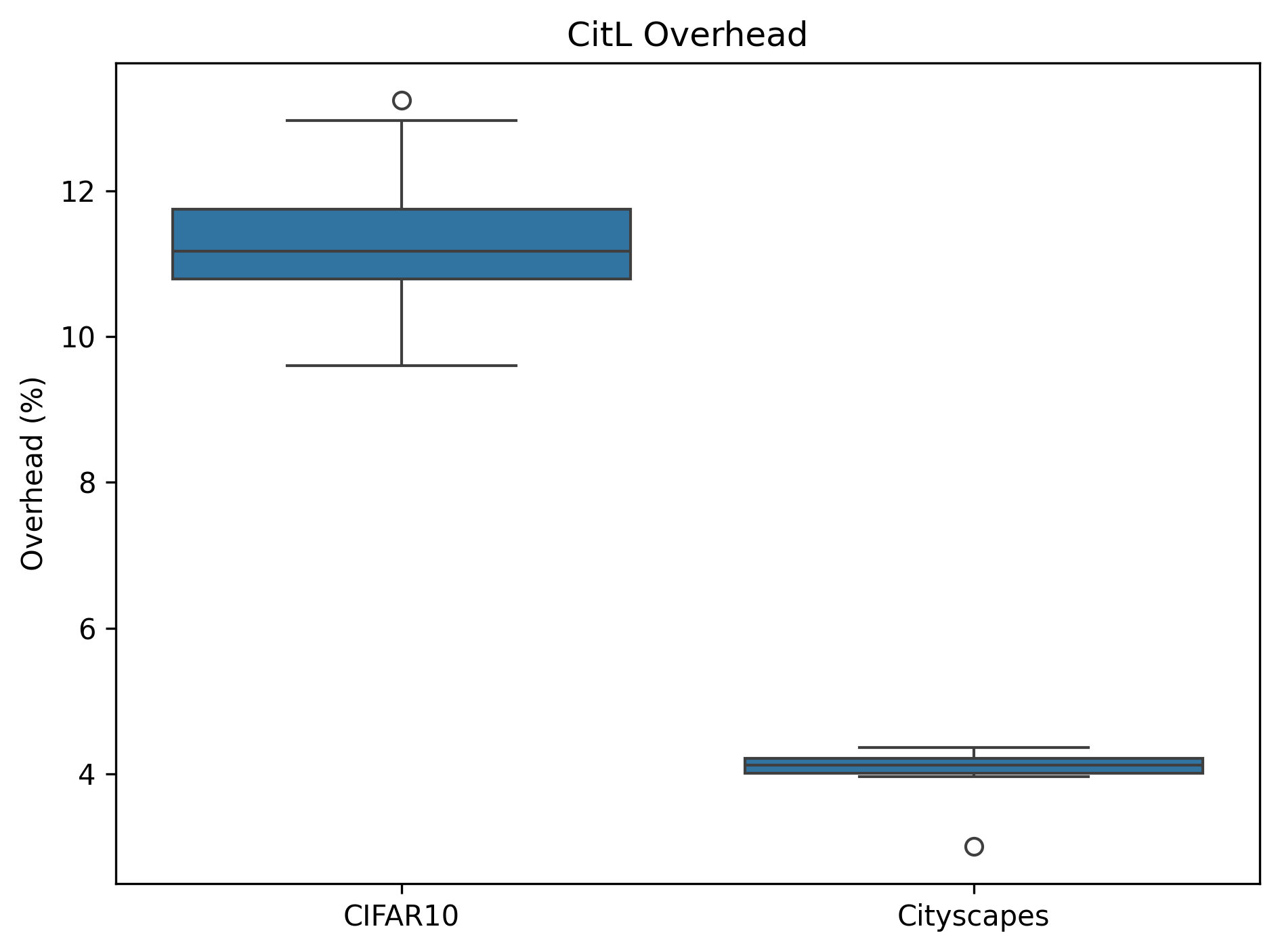}
    \caption{Average per training-step time of CitL over baseline.  Multiple CitL values represent different $\alpha$ hyperparameters. (CIFAR-10 11.2 $\pm$ 0.8\%, CityScapes 4.0 $\pm$ 0.4\%)}
    \label{fig:overhead}
    \end{minipage}\hfill
    \begin{minipage}{0.48\textwidth}
    \centering
    \small
    \captionof{table}{Percentage overhead of our CitL relative to the baseline, showing mean, standard deviation (Std), minimum (Min), and maximum (Max) values across datasets.}
    \label{tab:overhead}
    \begin{tabular}{lrrrr}
    \toprule
    dataset & mean & std & min & max \\
    \midrule
    CIFAR-10 & 11.2\% & 0.8 & 9.6 & 13.2 \\
    CityScapes & 4.0\% & 0.4 & 3.0 & 4.4 \\
    \bottomrule
    \end{tabular}
    \end{minipage}
\end{figure}

Fig.~\ref{fig:overhead} and Table~\ref{tab:overhead} present an analysis of the overhead associated with our method across various tasks. Notably, classification tasks demonstrate a higher overhead compared to segmentation tasks, likely because baseline training steps are faster. However, the overall increase in overhead remains small, with a mean increase of 11.2\% in training time per step for classification and 4.0\% for segmentation. These findings are encouraging and suggest that our method is likely to scale effectively to larger tasks and datasets.
This section presents an analysis of the weighting generated by our method on the imbalanced CIFAR-10 dataset and the CityScapes dataset. The primary objective is to demonstrate the method's effectiveness in addressing dataset imbalances by emphasizing challenging examples. 

\subsection{Example Weighting}

\begin{figure}[t]
    \centering
    \begin{subfigure}[b]{0.45\textwidth}   
        \includegraphics[width=\textwidth]{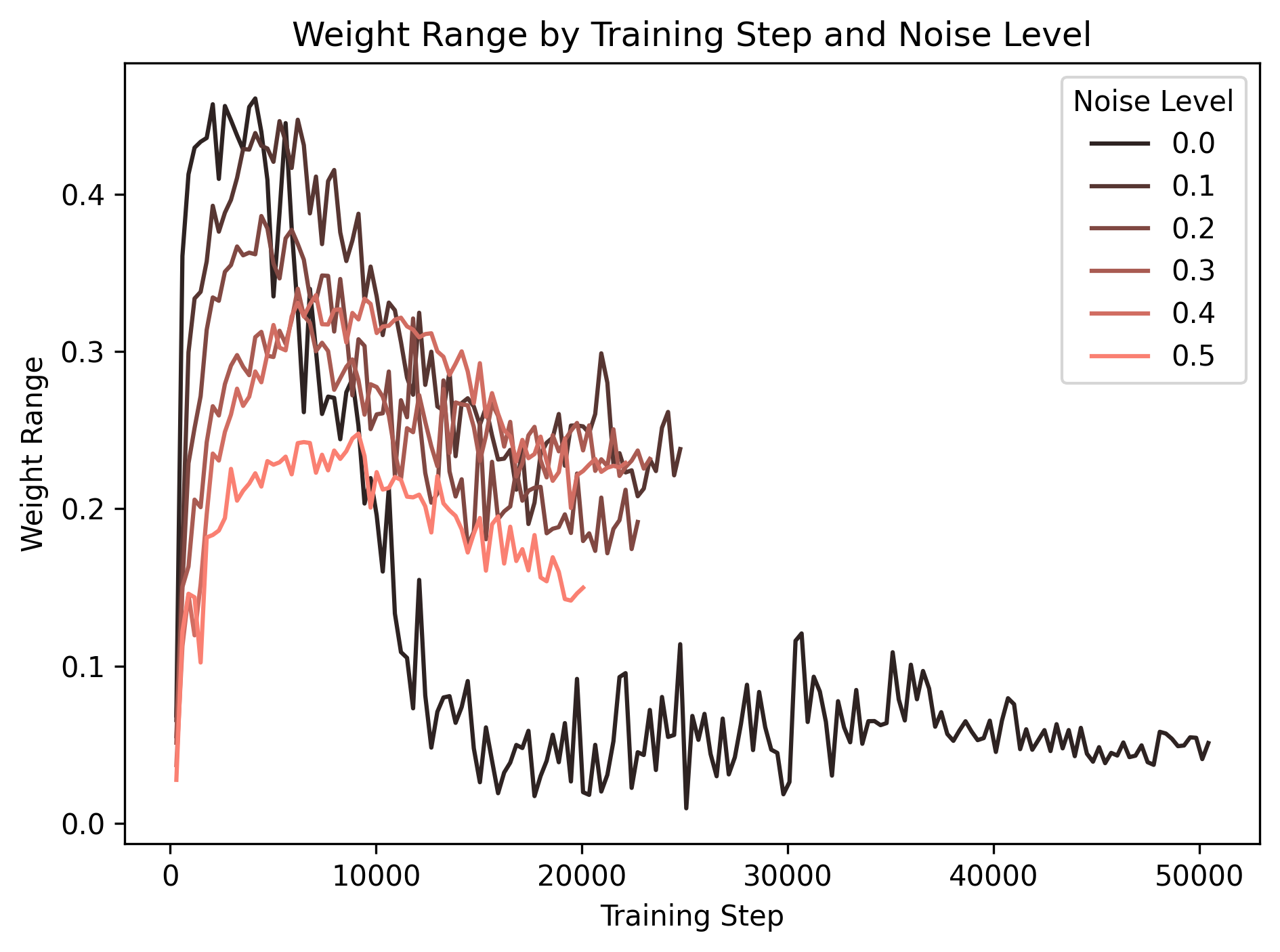}
        \caption{Weight Range, showing the difference in mean weights between the highest-weighted and lowest-weighted classes on the imbalanced CIFAR-10 dataset, stratified by noise level. Only the best performing $\alpha$ value is shown for each noise level.}
        \label{fig:weight_range}
    \end{subfigure}\hfill
    \begin{subfigure}[b]{0.45\textwidth}   
        \includegraphics[width=\textwidth]{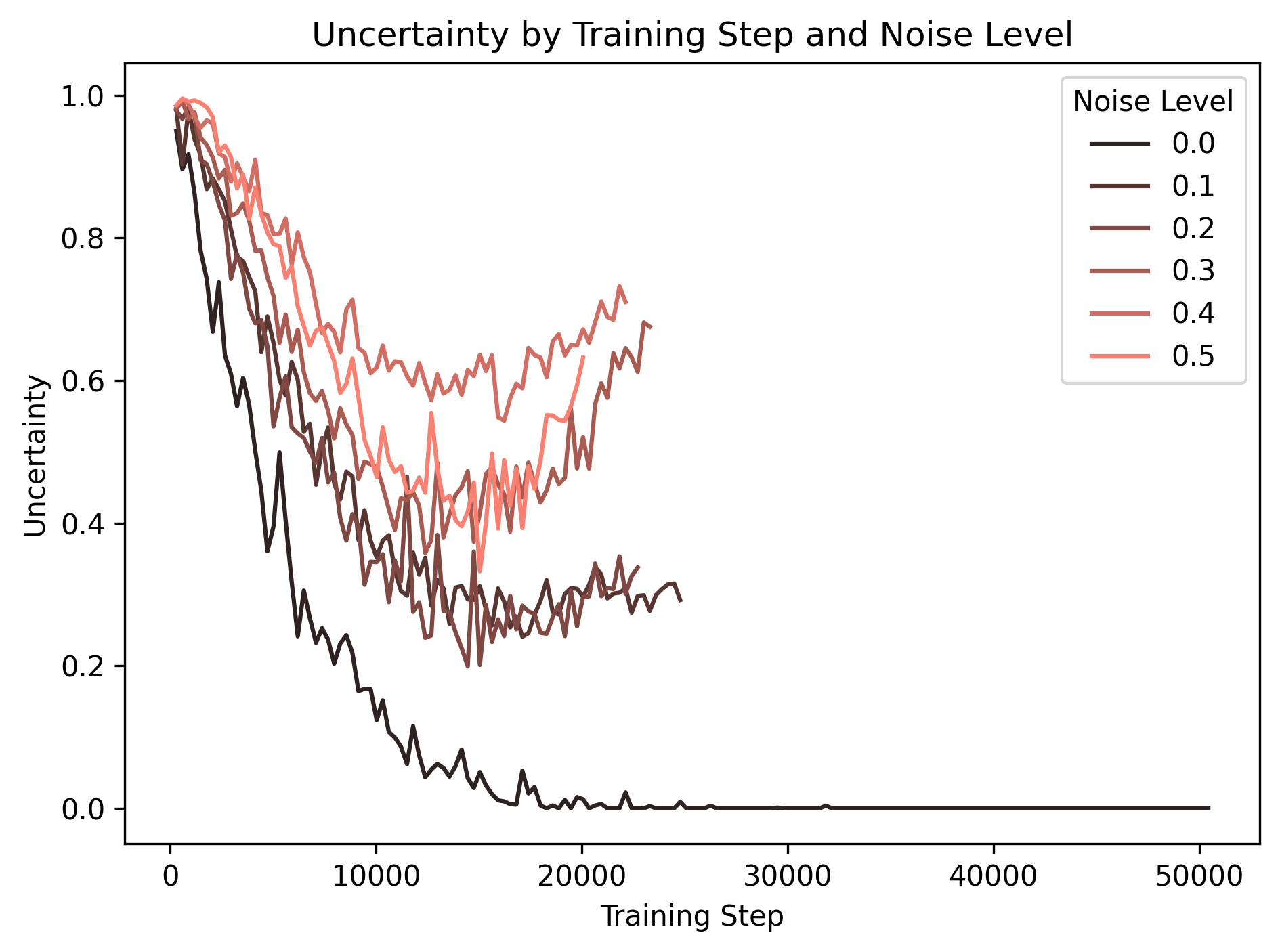}
        \caption{Uncertainty by noise level for the imbalanced CIFAR-10 dataset. Uncertainty is defined as the percentage of examples whose prediction set contains more than one possible class. With no noise, the model is able to reduce uncertainty to 0; noise makes this reduction to 0 impossible.}
        \label{fig:uncertainty}
    \end{subfigure}\hfill
\end{figure}

For the classification task, Fig.~\ref{fig:weight_range} illustrates the variation in mean weight between the highest- and lowest-weighted classes across different training steps. Notably, weighting is highest during the early training epochs and decreases as training progresses, reflecting the model's evolving uncertainty, as shown in Fig.~\ref{fig:uncertainty}. In the absence of noise, the model successfully reduces uncertainty to zero; however, this complete reduction is unattainable in the presence of noise. Similarly, noise inhibits the method's ability to diminish the weighting effect to zero.

\begin{figure}[p]
    \centering
    \begin{subfigure}[b]{0.45\textwidth}
        \centering
        \includegraphics[width=\textwidth, trim={0.8cm 0.7cm 0 0.8cm}, clip]{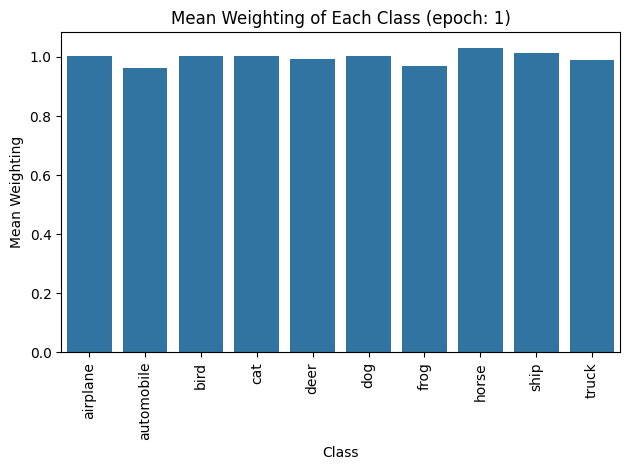}
        \caption{Epoch 1}
        \label{fig:cifar_epoch_1_weight}
    \end{subfigure}
    \hfill
    \begin{subfigure}[b]{0.45\textwidth}
        \centering
        \includegraphics[width=\textwidth, trim={0.8cm 0.7cm 0 0.8cm}, clip]{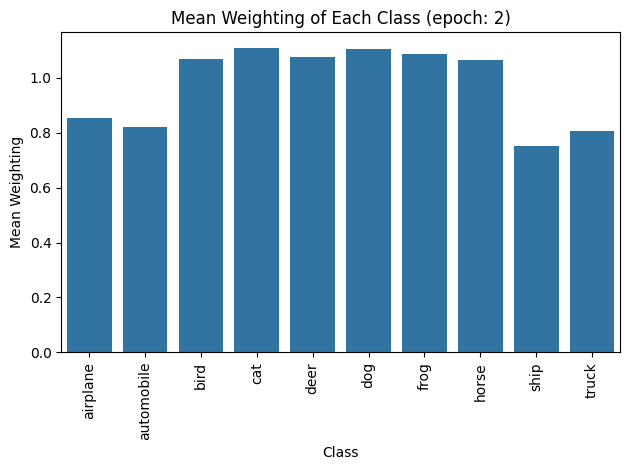}
        \caption{Epoch 2}
        \label{fig:cifar_epoch_2_weight}
    \end{subfigure}

    \vspace{0.1cm}
    
    \begin{subfigure}[b]{0.45\textwidth}
        \centering
        \includegraphics[width=\textwidth, trim={0.8cm 0.7cm 0 0.8cm}, clip]{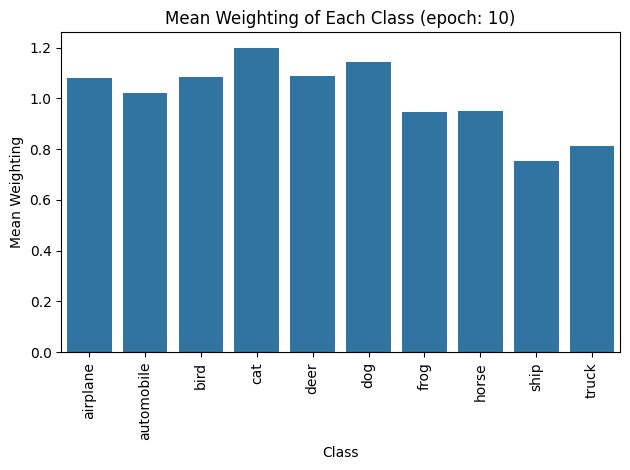}
        \caption{Epoch 10}
        \label{fig:cifar_epoch_10_weight}
    \end{subfigure}
    \hfill
    \begin{subfigure}[b]{0.45\textwidth}
        \centering
        \includegraphics[width=\textwidth, trim={0.8cm 0.7cm 0 0.8cm}, clip]{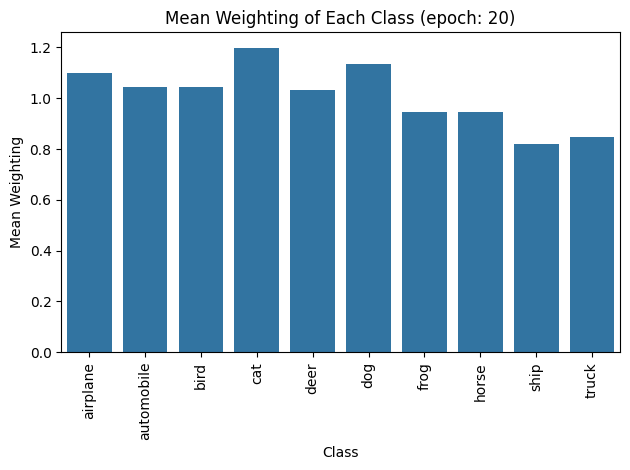}
        \caption{Epoch 20}
        \label{fig:cifar_epoch_20_weight}
    \end{subfigure}

    \vspace{0.1cm}
    
    \begin{subfigure}[b]{0.45\textwidth}
        \centering
        \includegraphics[width=\textwidth, trim={0.8cm 0.7cm 0 0.8cm}, clip]{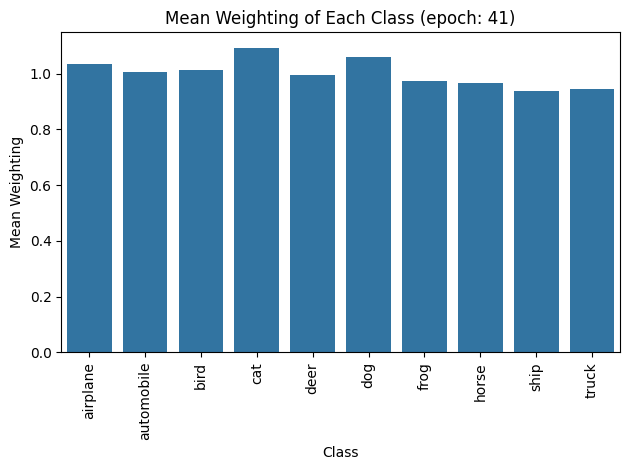}
        \caption{Epoch 40}
        \label{fig:cifar_epoch_40_weight}
    \end{subfigure}
    \hfill
    \begin{subfigure}[b]{0.45\textwidth}
        \centering
        \includegraphics[width=\textwidth, trim={0 0 0 0.8cm}, clip]{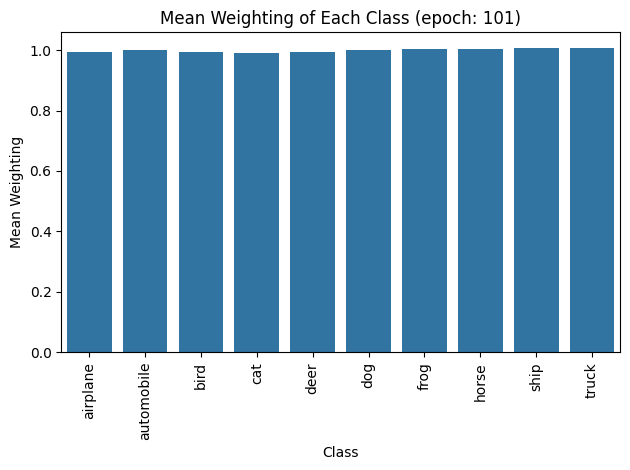}
        \caption{Epoch 100}
        \label{fig:cifar_epoch_100_weight}
    \end{subfigure}
        \caption{Mean weight by class across training epochs for the imbalanced CIFAR-10 dataset with no synthetic label noise. Each subfigure shows the distribution of mean class weights at different training epochs, illustrating how the algorithm adjusts class weights over time.}
    \label{fig:classification_mean_weights}
\end{figure}

\begin{figure}[p]
    \centering
    \begin{subfigure}[b]{0.45\textwidth}
        \centering
        \includegraphics[width=\textwidth, trim={0.8cm 0.7cm 0 0.8cm}, clip]{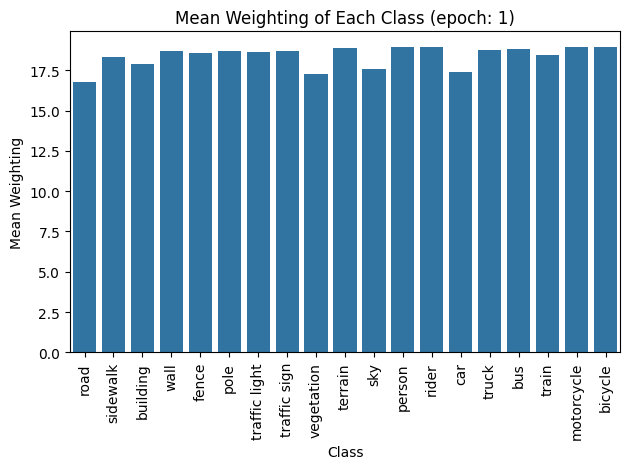}
        \caption{Epoch 1}
        \label{fig:epoch_1_weight}
    \end{subfigure}
    \hfill
    \begin{subfigure}[b]{0.45\textwidth}
        \centering
        \includegraphics[width=\textwidth, trim={0.8cm 0.7cm 0 0.8cm}, clip]{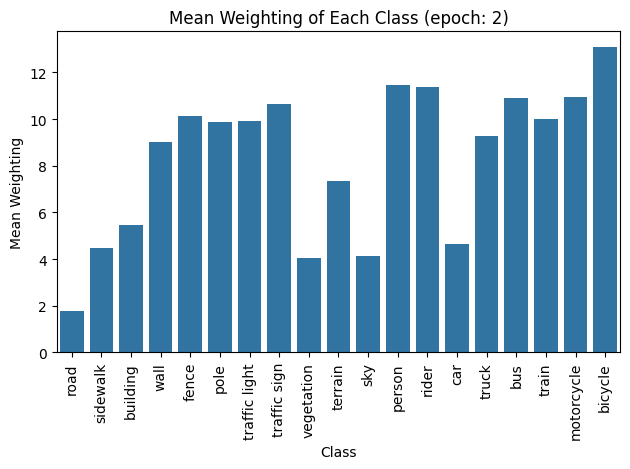}
        \caption{Epoch 2}
        \label{fig:epoch_2_weight}
    \end{subfigure}

    \vspace{0.1cm}
    
    \begin{subfigure}[b]{0.45\textwidth}
        \centering
        \includegraphics[width=\textwidth, trim={0.8cm 0.7cm 0 0.8cm}, clip]{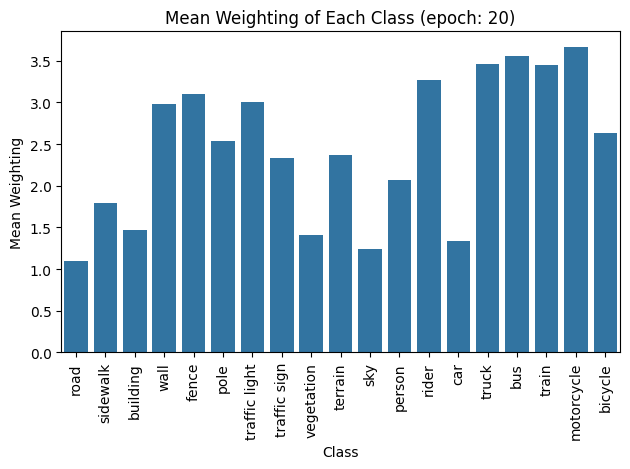}
        \caption{Epoch 20}
        \label{fig:epoch_20_weight}
    \end{subfigure}
    \hfill
    \begin{subfigure}[b]{0.45\textwidth}
        \centering
        \includegraphics[width=\textwidth, trim={0.8cm 0.7cm 0 0.8cm}, clip]{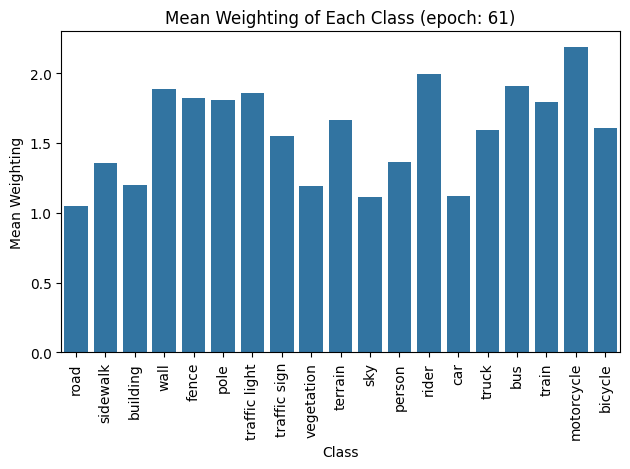}
        \caption{Epoch 60}
        \label{fig:epoch_60_weight}
    \end{subfigure}

    \vspace{0.1cm}
    
    \begin{subfigure}[b]{0.45\textwidth}
        \centering
        \includegraphics[width=\textwidth, trim={0.8cm 0.7cm 0 0.8cm}, clip]{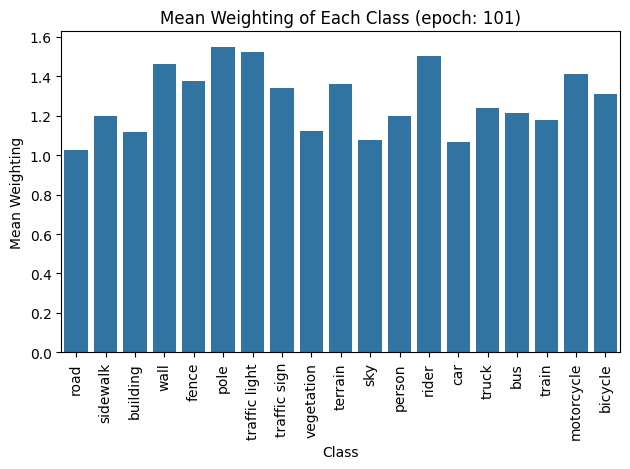}
        \caption{Epoch 100}
        \label{fig:epoch_100_weight}
    \end{subfigure}
    \hfill
    \begin{subfigure}[b]{0.45\textwidth}
        \centering
        \includegraphics[width=\textwidth, trim={0 0 0 0.70cm}, clip]{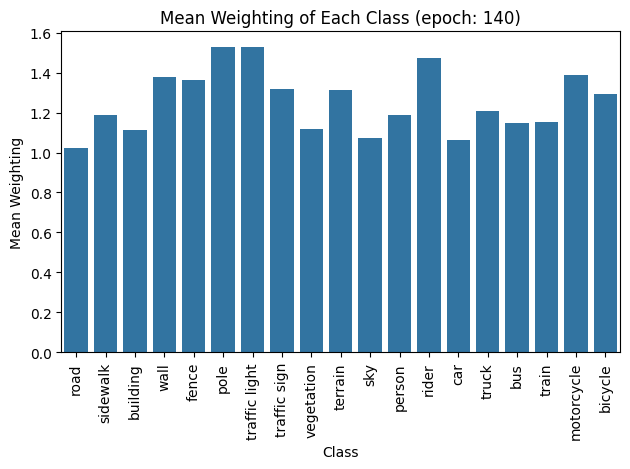}
        \caption{Epoch 140}
        \label{fig:epoch_140_weight}
    \end{subfigure}
        \caption{Mean weight by class across training epochs for the CityScapes dataset for best value of $\alpha$ = 0.02. Each subfigure shows the distribution of mean class weights at different training epochs, illustrating how the algorithm adjusts class weights over time.}
    \label{fig:segmentation_mean_weights}
\end{figure}

Fig.~\ref{fig:classification_mean_weights} illustrates the breakdown of weighting for CIFAR-10 by class throughout the training epochs. In Epoch 1, the network demonstrates a lack of learning about the dataset, assigning roughly equal weights to all classes. By Epoch 2, the model begins to produce lower weights for the ship and truck classes. Between Epochs 10 and 20, the method identifies the downsampled class of airplane as having a high level of uncertainty, thus assigning it a higher weight. In contrast, the other downsampled class, automobile, is not weighted as heavily, likely due to its shared features with other classes, allowing it to maintain good performance despite downsampling. Notably, the cat and dog classes receive the highest weights, likely due to their similarity and difficulty in distinction. As training progresses, the weighting for airplane, cat, and dog declines, reflecting the network's increasing certainty in its predictions, culminating in small weight adjustments by Epoch 40. By Epoch 100, the model shows negligible differences in class weighting.

Fig.~\ref{fig:segmentation_mean_weights} presents the breakdown of weighting for CityScapes by class. Initially, the model exhibits near-uniform uncertainty across classes. By Epoch 2, significant learning progress is evident, with easier classes such as road, vegetation, sky, and car attaining lower weights, while more challenging classes like bicycle receive higher weights. This trend continues as the model refines its weights over time; by Epoch 20, the average weight for all classes falls below 3. Notably, by Epoch 30, the weight for the motorcycle class rises to over 3.5, indicating the model's increased focus on harder-to-predict classes, despite overall improved performance. In Epochs 40 and 60, the model demonstrates reduced re-weighting, with only the most challenging classes (e.g., traffic light, traffic sign, rider, motorcycle, and bicycle) receiving weights near or exceeding 2.

\subsection{Example Pruning}
\label{sec:pruning}

\begin{figure}[t]
    \centering
    \begin{subfigure}{0.40\textwidth}
        \centering
        \includegraphics[width=\textwidth]{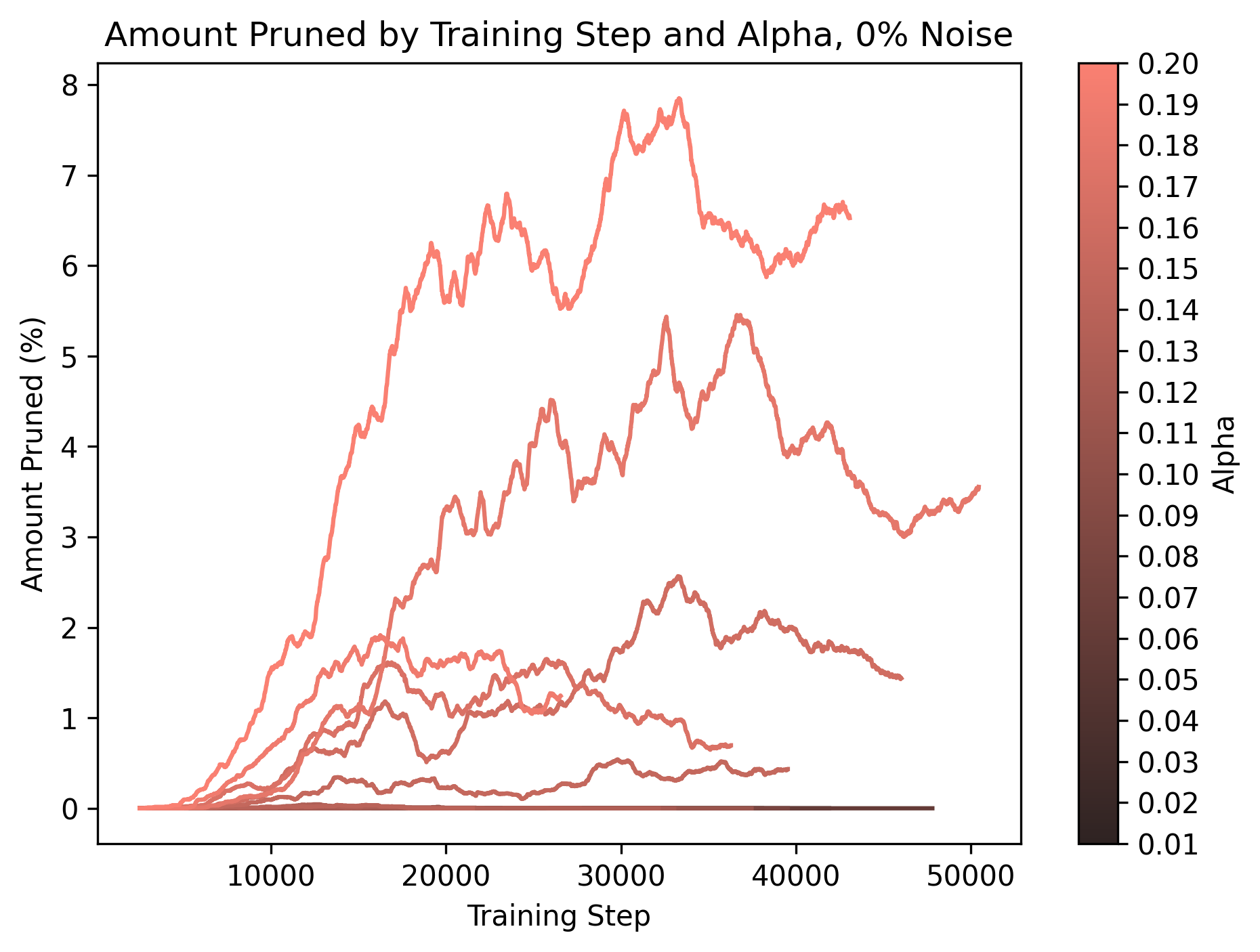}
        \caption{Imbalanced CIFAR-10 at noise level 0\%}
        \label{fig:pruned_noise_level_0}
    \end{subfigure}
    \hfill
    \begin{subfigure}{0.40\textwidth}
        \centering
        \includegraphics[width=\textwidth]{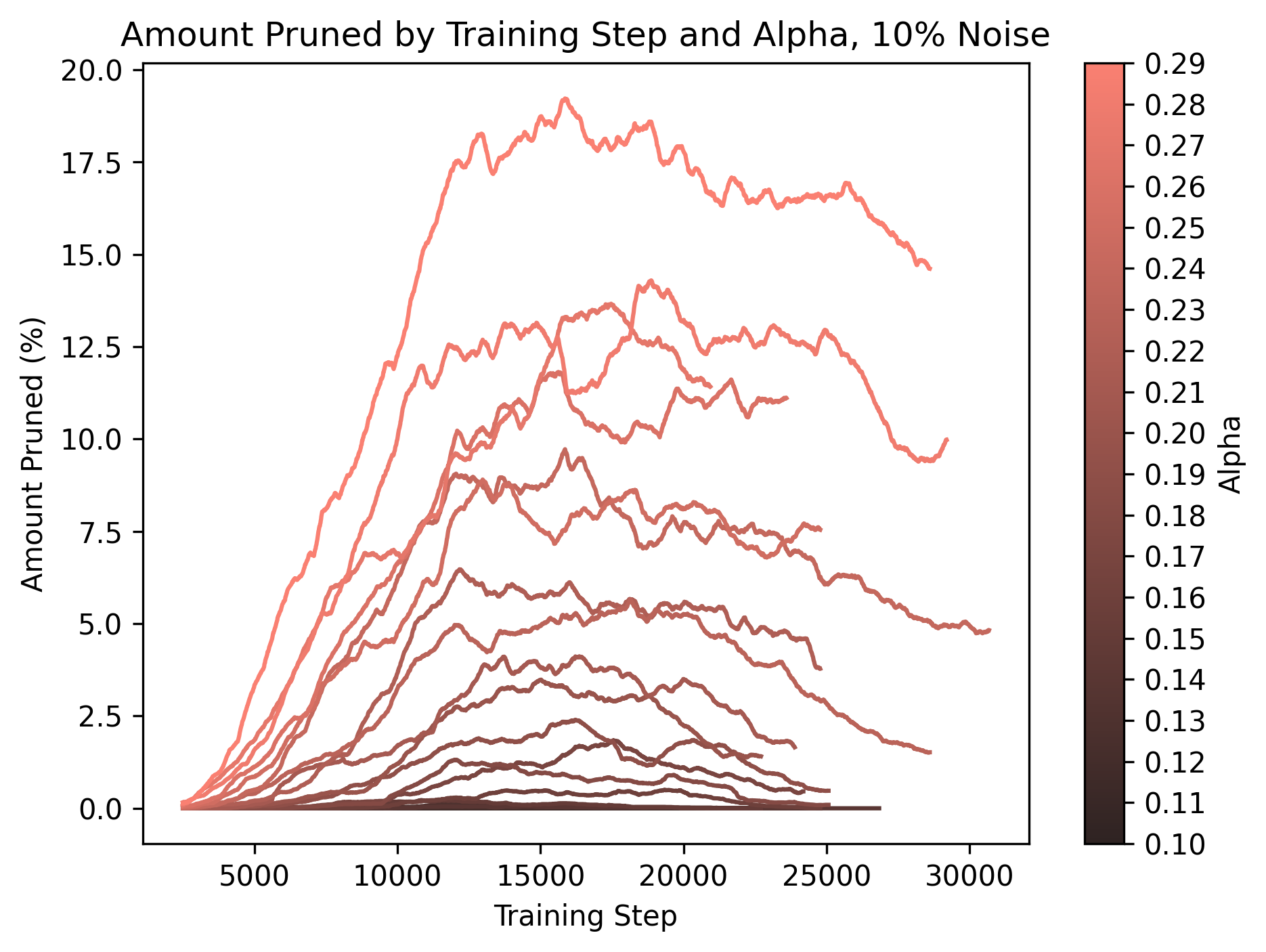}
        \caption{Imbalanced CIFAR-10 at noise level 10\% }
        \label{fig:pruned_noise_level_10}
    \end{subfigure}
    \hfill
    \begin{subfigure}{0.40\textwidth}
        \centering
        \includegraphics[width=\textwidth]{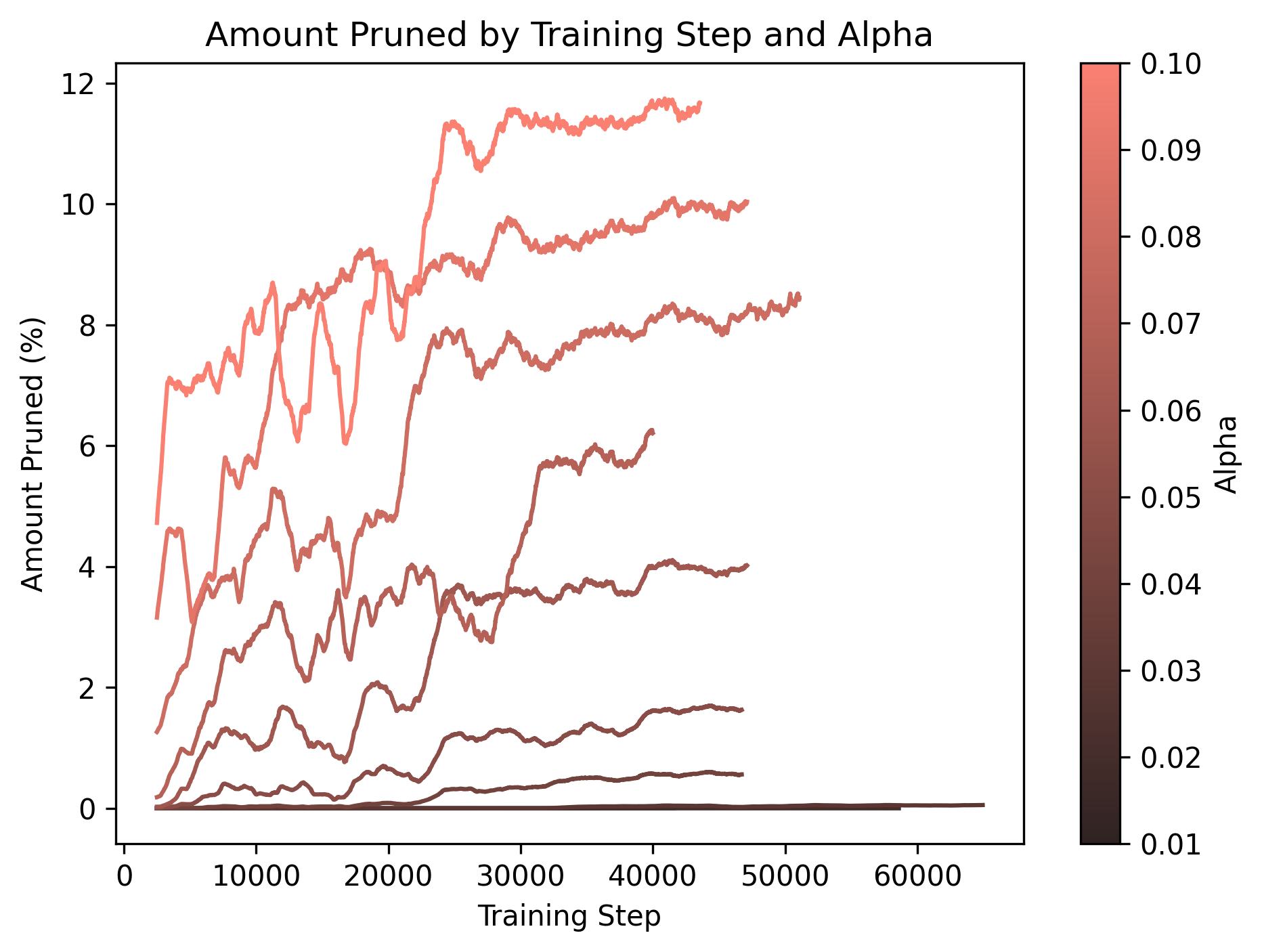}
        \caption{CityScapes }
        \label{fig:pruned_cityscapes}
    \end{subfigure}
    \caption{Amount Pruned by Training Step and Alpha.}
    \label{fig:pruned_amount}
\end{figure}

\begin{figure}[t]
    \centering
    \begin{subfigure}{0.40\textwidth}
        \centering
        \includegraphics[width=\textwidth]{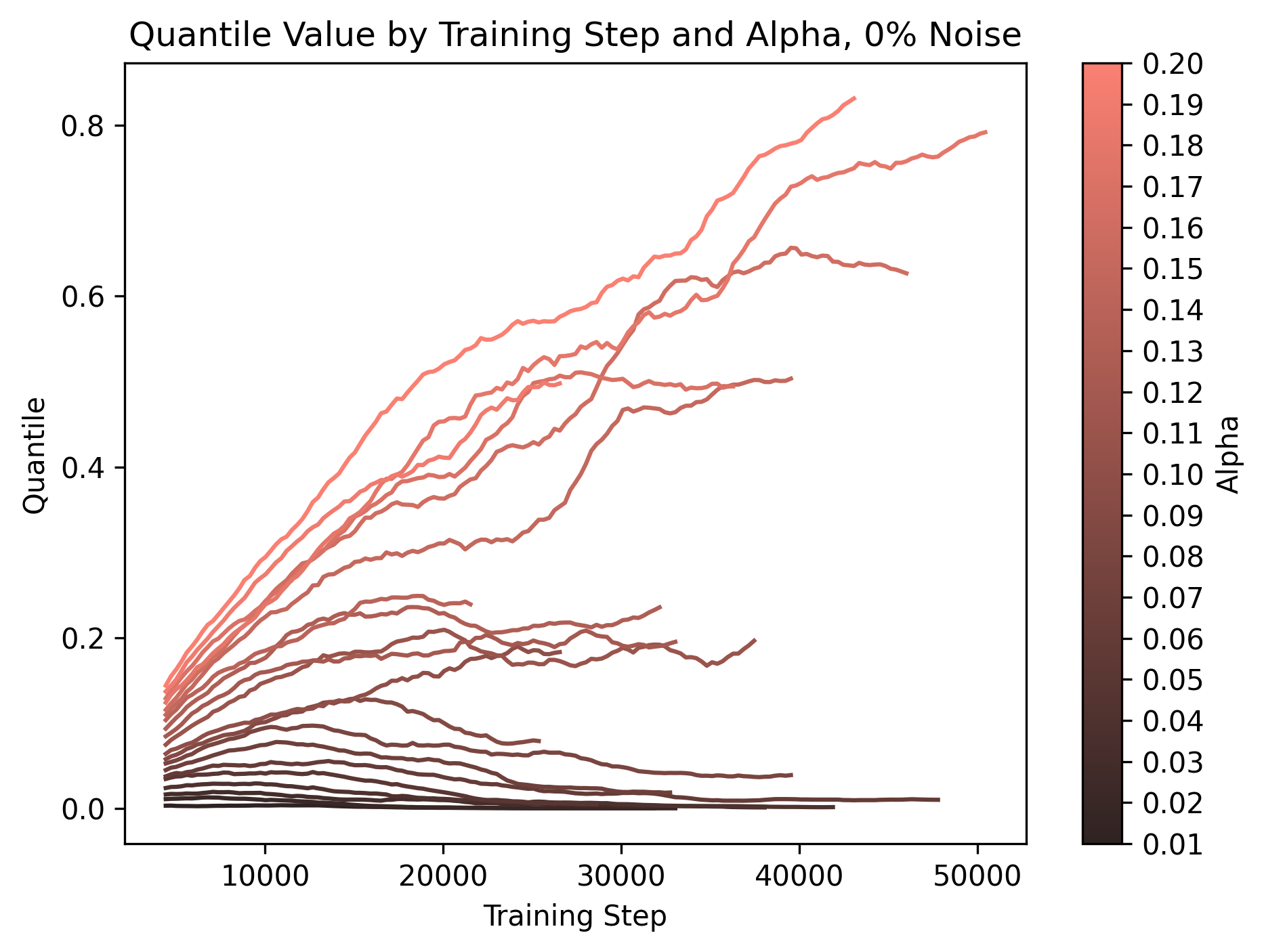}
        \caption{Imbalanced CIFAR-10 at noise level 0\%}
        \label{fig:quantile_noise_level_0}
    \end{subfigure}
    \hfill
    \begin{subfigure}{0.40\textwidth}
        \centering
        \includegraphics[width=\textwidth]{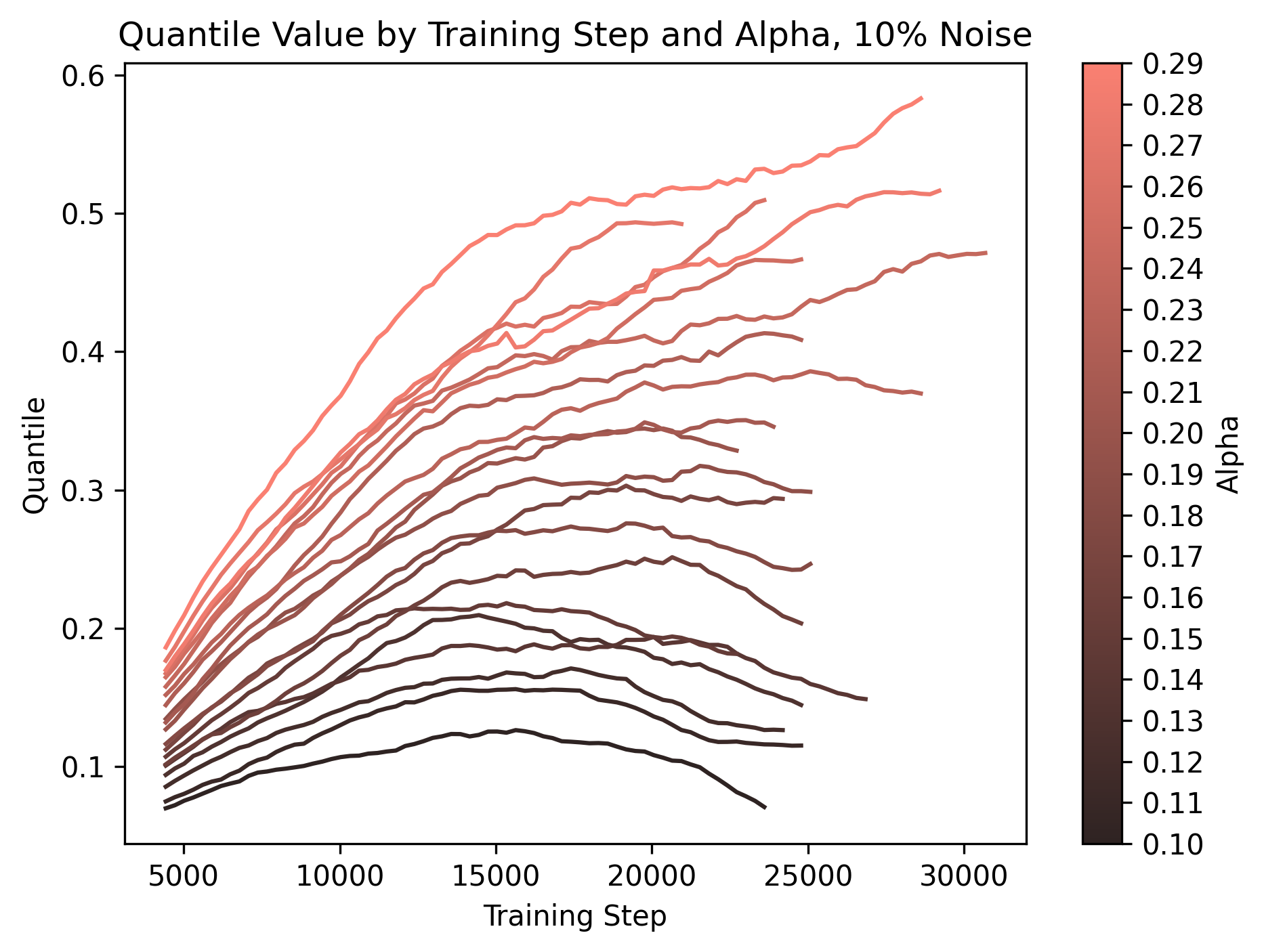}
        \caption{Imbalanced CIFAR-10 at noise level 10\% }
        \label{fig:quantile_noise_level_10}
    \end{subfigure}
    \hfill
    \begin{subfigure}{0.40\textwidth}
        \centering
        \includegraphics[width=\textwidth]{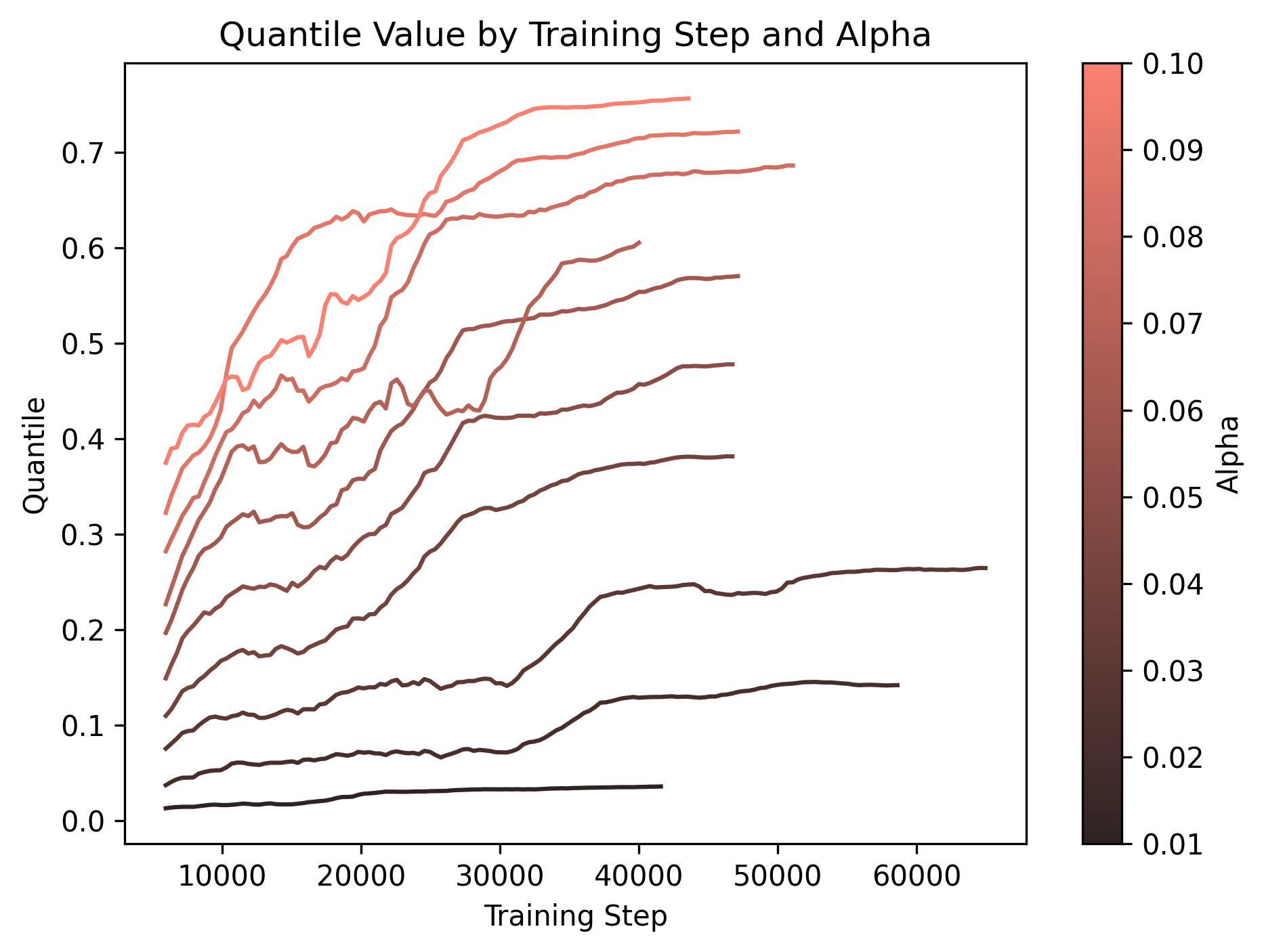}
        \caption{CityScapes }
        \label{fig:quantile_noise_level_cityscapes}
    \end{subfigure}
    \caption{Quantile by Training Step and Alpha.}
    \label{fig:noise_level}
\end{figure}

Pruning is controlled by $\alpha$, with higher values leading to more pruning.  From Fig.~\ref{fig:pruned_amount}, we can see that many of the lower values of $\alpha$ perform no pruning at all, relying entirely on weighting to balance classes.  Fig.~\ref{fig:pruned_noise_level_0} shows the percentage of examples pruned for the imbalanced CIFAR-10 dataset with no noise.  No pruning is done until $\alpha \ge 0.16$, and at the best $\alpha = 0.18$ approximately 2\% of the dataset is pruned.  Similarly for CityScapes, at the best $\alpha = 0.02$, approximately 1\% of the data is pruned.

\begin{figure}[t]
    \centering
    \begin{subfigure}[b]{0.45\textwidth}
        \centering
        \includegraphics[width=\textwidth, trim={0.8cm 0.7cm 0 0.8cm}, clip]{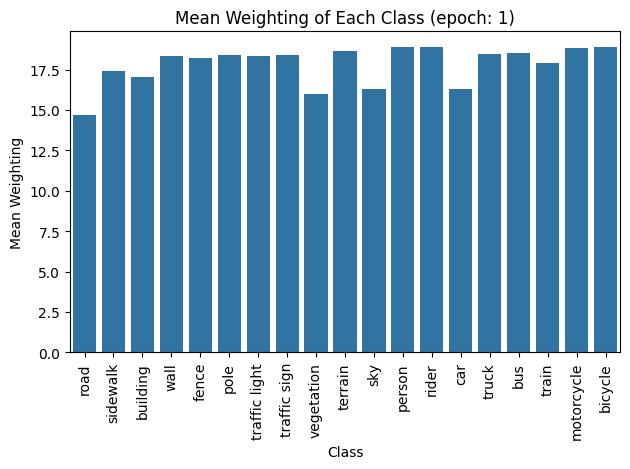}
        \caption{Epoch 1}
        \label{fig:pruned_epoch_1_weight}
    \end{subfigure}\hfill
    \begin{subfigure}[b]{0.45\textwidth}
        \centering
        \includegraphics[width=\textwidth, trim={0.8cm 0.7cm 0 0.8cm}, clip]{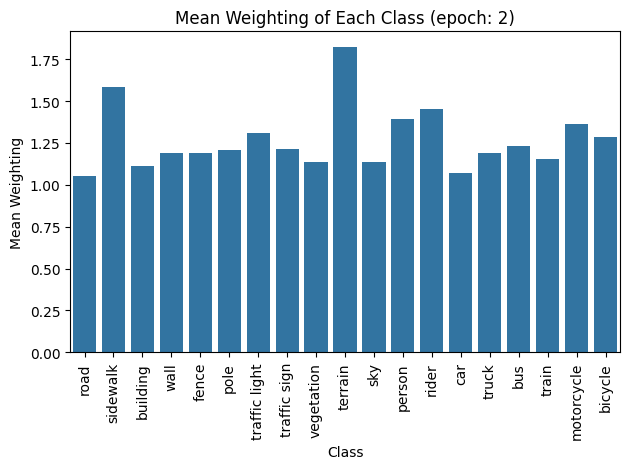}
        \caption{Epoch 2}
        \label{fig:pruned_epoch_2_weight}
    \end{subfigure}

    \vspace{0.1cm}
    
    \begin{subfigure}[b]{0.45\textwidth}
        \centering
        \includegraphics[width=\textwidth, trim={0.8cm 0.7cm 0 0.8cm}, clip]{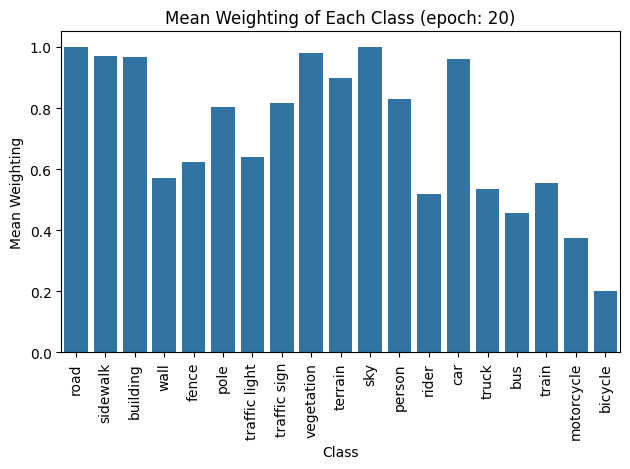}
        \caption{Epoch 20}
        \label{fig:pruned_epoch_20_weight}
    \end{subfigure}
    \hfill
    \begin{subfigure}[b]{0.45\textwidth}
        \centering
        \includegraphics[width=\textwidth, trim={0.8cm 0.7cm 0 0.8cm}, clip]{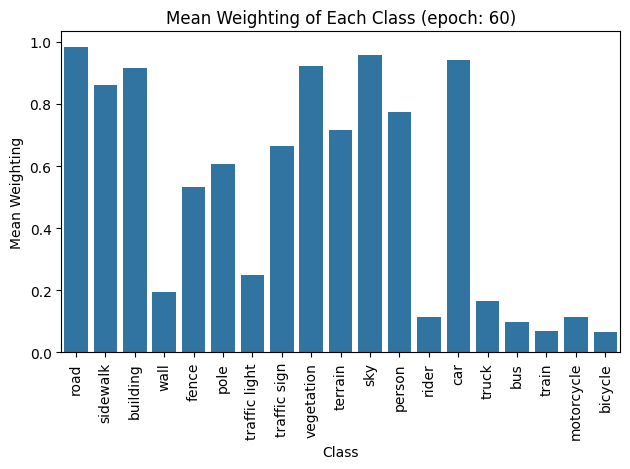}
        \caption{Epoch 60}
        \label{fig:pruned_epoch_60_weight}
    \end{subfigure}
    \caption{Mean weight by class across training epochs for the CityScapes dataset at worst value for $\alpha$ = 0.09. Each subfigure shows the distribution of mean class weights at different training epochs, illustrating how the algorithm adjusts class weights over time.}
    \label{fig:pruned_mean_weights}
\end{figure}

High values of $\alpha$ result in poor performance due to the preferential pruning of highly uncertain examples, which are usually from minority classes.  This effect is shown in Fig.~\ref{fig:pruned_mean_weights} for CityScapes at the worst $\alpha = 0.09$.  At epoch 2, the relatively easy classes of sidewalk and terrain are weighted very highly.  By epoch 20, the hard classes are heavily pruned; by epoch 60 the classes of wall, traffic light, rider, truck, bus, train, motorcycle and bicycle are nearly pruned completely.

\begin{figure}[t]
    \centering
    \includegraphics[width=0.42\textwidth]{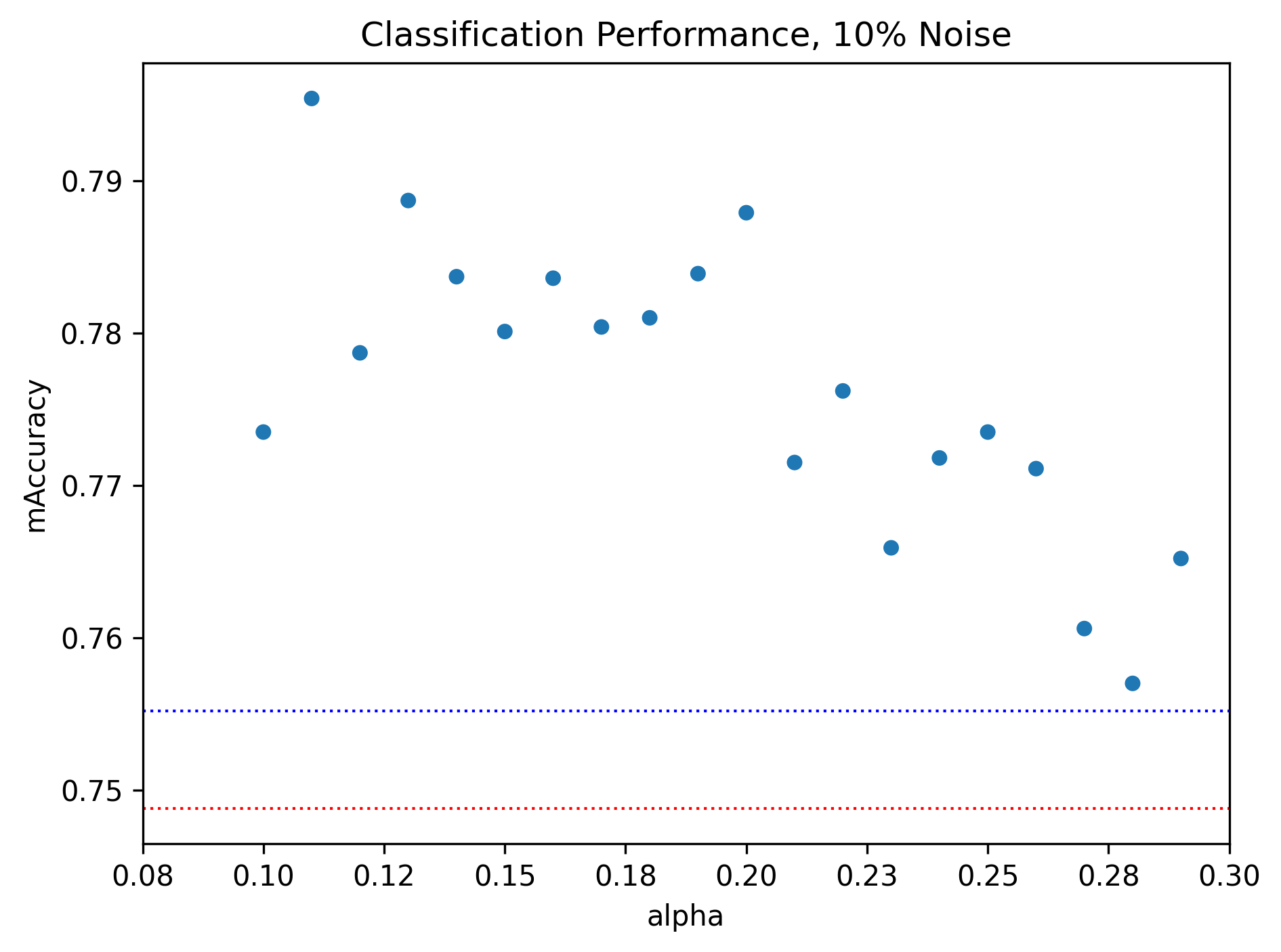}
    \caption{Imbalanced CIFAR-10 classification performance at 10\% noise level.  The blue dotted line represents the baseline using Cross Entropy Loss; the red dotted line represents the baseline using Focal Loss.  The method outperforms both baselines at all tested values of $\alpha$.  At $\alpha = 0.20$,  the method prunes $\approx$ 2.5\% of examples.  Performance declines when $\alpha \ge 0.21$.}
    \label{fig:cifar_perforamce_0.10}
    \vspace{1cm}
\end{figure}

The introduction of noise adds more uncertainty to the network which reduces the amount of pruning overall. Again, no pruning is done until $\alpha \ge 0.16$; moreover, the amount of pruning is reduced when compared to the 0\% noise experiment.  Pruning to the amount of noise (10\%) is not achieved until approximately $\alpha = 0.24$.  Fig.~\ref{fig:cifar_perforamce_0.10} shows that the method performs well until $\alpha \ge 0.21$, at which point it starts to degrade in the same fashion as CityScapes experiment explored in Fig.~\ref{fig:segmentation_results}. Still, for all values of $\alpha$ until 0.30, the method outperforms both the cross entropy loss and focal loss baselines.  At $\alpha = 0.20$, the last before the method begins to degrade, the method prunes approximately 2. 5\% of the training examples.

\newpage

\subsection{Considerations for Selecting $\alpha$}

\label{sec:selecting_alpha}

In the proposed method, the parameter \(\alpha\) serves as a crucial balancing factor between the weighting of uncertain examples and the pruning of extremely challenging examples. Lower values of \(\alpha\) favor increased weighting and reduced pruning, whereas higher values promote greater pruning and diminished weighting. The significance of the 0.50 quantile cannot be overstated, as it represents the threshold at which the maximum prediction set size is limited to 1. At this threshold, the network can influence relative class weights solely through pruning.

In the context of the CityScapes dataset, this critical threshold is reached with \(\alpha\) values of 0.06 and above (see Fig.~\ref{fig:quantile_noise_level_cityscapes}). Notably, this is the point where the method begins to underperform compared to the baseline (see Fig.~\ref{fig:segmentation_results}), indicating that \(\alpha\) should be chosen to maintain the quantile below this level when dealing with well-labeled, large datasets.

Low values of \(\alpha\) can lead the algorithm to overemphasize hard examples, which may decrease the model's certainty over time (see Fig.~\ref{fig:uncertainty}). This phenomenon is reflected in the quantile value, which one would expect to increase as the network undergoes training. In Fig.~\ref{fig:quantile_noise_level_0}, for \(\alpha\) values between 0.10 and 0.15, the quantile appears to plateau; moreover, for \(\alpha\) values below 0.10, the quantile declines towards the end of training. This decline is likely a result of the network overfitting to uncertain examples. The impact of this effect becomes more pronounced with the introduction of noise, necessitating even higher values of \(\alpha\) to sustain an increasing quantile (see Fig.~\ref{fig:quantile_noise_level_10}).

For the CityScapes dataset, the quantile increases across all values of \(\alpha\). We attribute this to the high quality of labeling in this finely labeled dataset, coupled with the significant number of examples present in the segmentation task. These factors contribute to a smoother loss landscape. Consequently, our method may be better suited for datasets with a large number of examples per batch, suggesting its greater efficacy in segmentation tasks compared to classification.

\section{Conclusions}
In this paper, we propose a novel method, Conformal-in-the-Loop (CitL), which integrates Conformal Prediction (ConP) directly into the training process to create an uncertainty-aware approach for weighting and pruning training examples with minimal computational overhead. Our approach shows improved accuracy in noisy, imbalanced multiclass classification on the CIFAR-10 dataset and achieves superior mIoU on the CityScapes dataset. CitL is particularly effective in addressing small, underrepresented classes that are often overshadowed by dominant, easily segmented classes. We also present an in-depth analysis of CitL's mechanisms, demonstrating how it prioritizes uncertain examples early in training and increasingly weights marginal examples later, highlighting its adaptability. Notably, this is accomplished using a single hyperparameter, $\alpha$, without requiring prior knowledge of the dataset structure.

The main limitation of this work is the challenge of distinguishing mislabeled examples from unlearned examples in very noisy datasets (greater than 10\%). This difficulty can result in preferential pruning of the minority class. Confident Learning addresses this issue by utilizing class-conditional noise rates~\cite{northcutt2021confident}. An intriguing extension of our work would be to apply class-conditional conformal prediction~\cite{NEURIPS2023_cb931edd}.

The weighting of minority and majority classes is governed by the hyperparameter $\alpha$, which adjusts the size of the prediction set. However, the size of this prediction set depends on the number of classes in the dataset. Consequently, for smaller tasks, such as binary classification, the maximum weighting is also relatively low. While this is intuitively reasonable—since fewer classes can simplify the problem—it would be worthwhile to explore alternative weighting schemes, such as a non-linear function of the prediction set size.

Using LAC to manage both weighting and pruning introduces a discontinuity for highly uncertain examples when they transition from maximum weighting to being pruned. Although this behavior is comparable to other loss weighting methods~\cite{Jiang2018MentorNetLearningData}, our findings indicate that the primary advantage of our method stems from the weighting process. It may be worthwhile to explore alternative conformal prediction methods, such as APS~\cite{romano2020classification}, to enable independent weighting and pruning.

The simultaneous control of pruning and weighting creates sensitivity to the hyperparameter $\alpha$, which we addressed through a grid search. Future work could explore dynamically determining $\alpha$ using heuristic methods. This exploration could build on the insights from Section~\ref{sec:selecting_alpha}, which suggests maintaining the quantile below 0.5 while avoiding quantile collapse due to excessive focus on uncertain examples.

In conclusion, our method introduces a pruning and weighting mechanism that effectively addresses class imbalance and label noise, demonstrating superior performance at the lower noise levels typically encountered in practice. We evaluate CitL across multiclass classification and semantic segmentation tasks, highlighting performance enhancements in both domains. Our results indicate that CitL consistently improves model performance, achieving up to a 6.1\% increase in classification accuracy and a 5.0 mIoU improvement in segmentation. Moreover, the computational overhead is minimal, with training time per step increasing by only 11\% for classification tasks and 4\% for segmentation tasks.

\section*{Acknowledgements}
We acknowledge the support of the Natural Sciences and Engineering Research Council of Canada (NSERC), [funding reference number RGPIN-2023-05408].

Cette recherche a été financée par le Conseil de recherches en sciences naturelles et en génie du Canada (CRSNG), [numéro de référence RGPIN-2023-05408].

\bibliographystyle{splncs04}
\bibliography{refs}

\end{document}